\documentclass[10pt,twocolumn,letterpaper]{article}
\usepackage{cvpr}            
\usepackage{makecell}
\usepackage{multirow}
\usepackage{float}
\usepackage{mathbbol}
\usepackage{comment}
\usepackage{pifont}
\usepackage{tabularx}
\usepackage{supertabular}
\usepackage{wrapfig}
\usepackage{diagbox}
\usepackage{algorithm}
\usepackage{algpseudocode}
\usepackage{amsmath} 
\usepackage{graphicx}
\usepackage{booktabs} 
\usepackage{amssymb}
\usepackage{adjustbox}
\usepackage[dvipsnames]{xcolor}

\newcommand{\cmark}{\ding{51}}
\newcommand{\xmark}{\ding{55}}
\newcommand{\M}{\mathcal{M}}
\newcommand{\N}{\mathcal{N}}
\newcommand{\T}{\mathcal{T}}
\DeclareMathOperator{\sinkhorn}{sinkhorn}
\DeclareMathOperator{\cosim}{cos\text{-}sim}
\DeclareMathOperator{\mcubes}{m\text{-}cubes}
\DeclareMathOperator{\apdef}{apply\text{-}def}
\def\eg{\emph{e.g.}}

\def\etal{\emph{et al.}}
\def\ie{\emph{i.e.}}
\def\wrt{\emph{w.r.t }}
\definecolor{cvprblue}{rgb}{0.21,0.49,0.74}
\usepackage[pagebackref,breaklinks,colorlinks,citecolor=cvprblue]{hyperref}

\title{Matching Shapes Under Different Topologies: A Topology-Adaptive Deformation Guided Approach}

\author{$^{1}$Aymen Merrouche
\quad
$^{2}$Stefanie Wuhrer
\quad
$^{2}$Edmond Boyer\\
Univ. Grenoble Alpes, CNRS, Inria, Grenoble INP, LJK, France \\
{\tt\small $^{1}$aymerrouche@gmail.com,  $^{2}$name.surname@inria.fr}
}

\begin{document}

\maketitle

\begin{abstract}
    Non-rigid 3D mesh matching is a critical step in computer vision and computer graphics pipelines. We tackle matching meshes that contain topological artefacts which can break the assumption made by current approaches. While Functional Maps assume the deformation induced by the ground truth correspondences to be near-isometric, ARAP-like deformation-guided approaches assume the latter to be ARAP. Neither assumption holds in certain topological configurations of the input shapes. We are motivated by real-world scenarios such as per-frame multi-view reconstructions, often suffering from topological artefacts.
    To this end, we propose a topology-adaptive deformation model allowing changes in shape topology to align shape pairs under ARAP and bijective association constraints.
    Using this model, we jointly optimise for a template mesh with adequate topology and for its alignment with the shapes to be matched to extract correspondences. We show that, while not relying on any data-driven prior, our approach applies to highly non-isometric shapes and shapes with topological artefacts, including noisy per-frame multi-view reconstructions, even outperforming methods trained on large datasets in 3D alignment quality. 
\end{abstract}

\section{Introduction}
\begin{figure}[h]
    \centering
    \includegraphics[width=0.8\columnwidth]{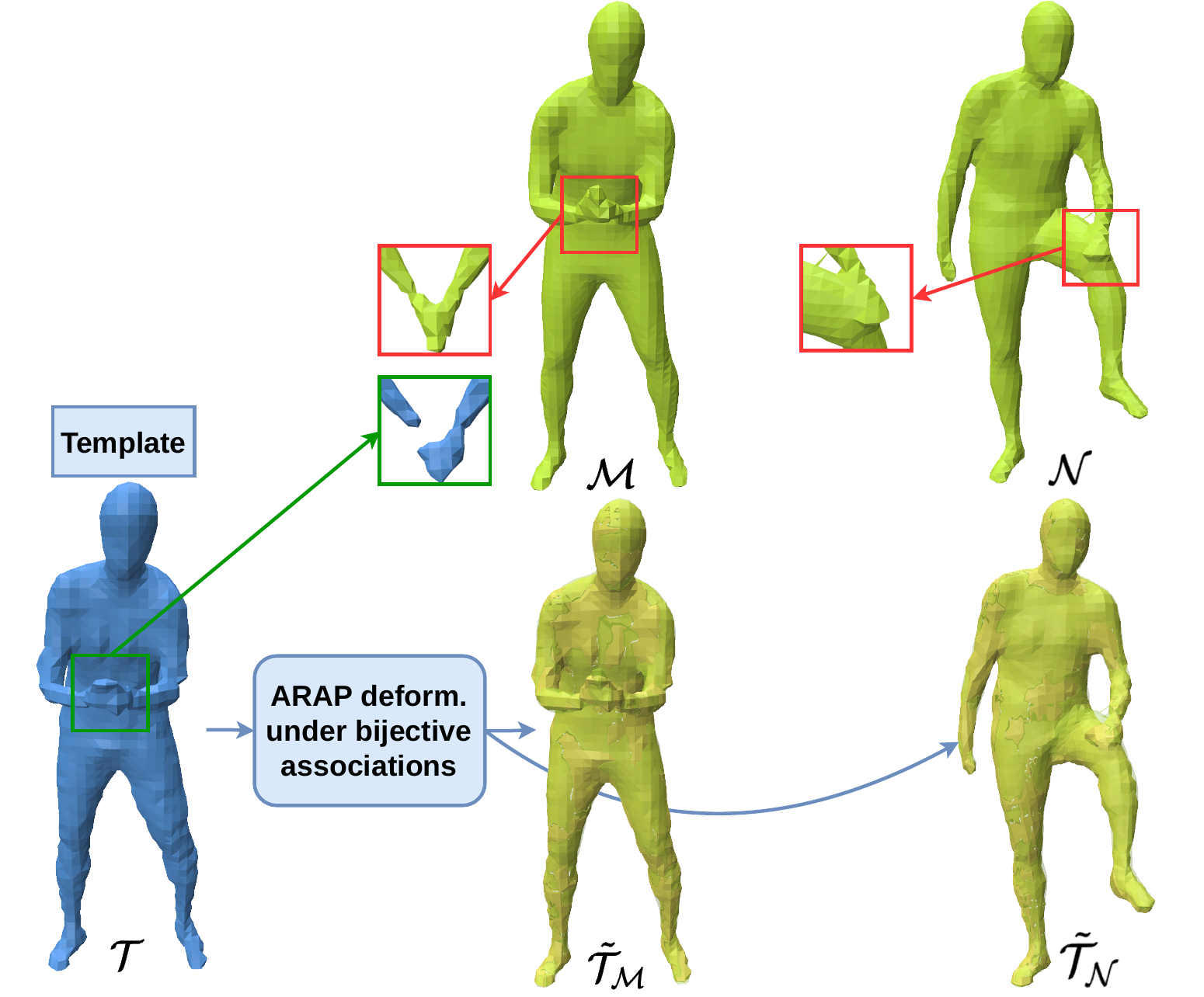}
    \caption{To match two meshes $\M$ and $\N$ with topological artefacts (see red boxes), we jointly optimise for a template $\T$, initialised as either $\M$ or $\N$, and its alignments $\tilde{\T}_{\M}$ and $\tilde{\T}_{\N}$ with $\M$ and $\N$ using a novel topology adaptive deformation model (see green box) under ARAP and bijective associations constraints.}
    \label{fig:teaser}
\end{figure}

Non-rigid 3D shape matching is the problem of finding dense correspondences between non-rigidly deforming 3D shapes. It establishes maps between shapes, which are essential for many 3D shape processing pipelines, including deformation transfer~\eg~\cite{basset2021neural}, motion re-targeting~\eg~\cite{Aberman2020, Lim2020} and statistical shape modeling~\eg~\cite{Pishchulin2017}.

In this work, we explicitly address the problem of matching 3D meshes with topological artefacts. Here, mesh topology refers to the topology of the underlying 2D manifold represented by the mesh. Variations in triangulation, sampling, or mesh density over the same manifold are considered changes in discretization.

We are motivated by real-world scenarios such as per-frame multi-view reconstructions, which, despite achieving strong geometric accuracy, often suffer from topological artifacts \eg~ close surface parts glued together. This distorts the intrinsic geometry of the shapes and pushes current approaches based on Functional Maps (FM)~\eg~\cite{litany2017deep,li2022attentivefmaps} or as-rigid-as-possible (ARAP) like~\cite{sorkine2007rigid} deformation models~\eg~\cite{eisenberger2021neuromorph, Merrouche_2023_BMVC} or both~\eg~\cite{eisenberger2020deep, cao2024spectral}, to their limits, restricting their applicability in real-world scenarios. 

While FMs assume a near-isometric deformation due to the reliance on the eigenfunctions of the Laplace Bletrami Operator which are only stable under near-isometries, ARAP like deformation models assume ARAP deformations leveraging the mesh connectivity to define rigidity constraints. For example, the deformation induced by the ground truth correspondences between the two meshes in the top row of Fig.~\ref{fig:teaser} is neither near-isometric nor ARAP.

On the other hand, point cloud matching approaches~\eg~\cite{zeng2021corrnet3d, lang2021dpc} fully discard the shape topology, falling short of meeting the performance of their mesh-based counterparts due to the lack of intrinsic information. As topological artifacts often concern limited shape parts, we propose a novel approach that adaptively adjusts the shape's topology when it impedes the matching objective.

We propose a topology-adaptive deformation model that allows changes in shape topology to align mesh pairs while enforcing as-rigid-as-possible and bijective associations constraints. Building on this, we design a mesh matching strategy jointly optimizing for a template mesh with adequate topology along with its alignment to the target shapes (see Fig.~\ref{fig:teaser}). While one could consider using a fixed, pre-defined template, identifying a universal template that works across diverse shape classes, such as clothed humans and animals, remains infeasible.

Our strategy initialises the template as either one of the two meshes to be matched and then iterates between optimising for an ARAP alignment under bijective associations with both meshes and editing the template shape's topology where this alignment is not feasible.  

A patch-based mesh representation is used both to define associations as a small matrix subject to bijectivity constraints and to efficiently parameterise the ARAP deformation they induce. Topology editing steps, are carried in a volumetric neural signed distance representation where a mesh's topology can be updated by editing this representation and then using an iso-surface extraction algorithm.

We evaluate our approach on matching human and animal shapes and demonstrate its robustness to non-isometric deformations and topological artefacts. We show its applicability on noisy per-frame multi-view reconstructions.

Our contributions are as follows: 1. A topology-adaptive deformation model to align meshes with topological artefacts under ARAP and bijective associations constraints 2. A topology-adaptive mesh matching strategy.

\section{Related Works}

Shape matching is a longstanding research subject. Early approaches rely on preserving handcrafted descriptors of points~\eg~\cite{Sun2009, Aubry2011, Zaharescu2009, Litman2011, Salti2014} or pairs of points~\eg~\cite{EladElbaz2003, Bronstein2006, Vestner2017, Coifman2005, Bronstein2010, Vestner2018}. Rodolà~\etal~\cite{Rodol2014} proposed to use random forests to learn correspondences, followed by works that cast shape matching as a vertex labelling problem~\eg~\cite{Masci2016, Boscaini2016, Boscaini20162, Monti2017}. 

In this related works we focus on recent strategies and put them in two main lines of work. First, deformation guided approaches, dominated by deep learning strategies, that find correspondences by aligning shapes with a deformation in 3D space. Second, Functional Maps (FM) that match shapes within the spectrum of some geometric operator spanning both learning and optimisation paradigms.  We refer the readers to surveys for both traditional~\cite{vanKaick2011, Biasotti2016} and recent methods~\cite{Deng2022} for an exhaustive literature review.

 In this paper, we present an optimisation-based deformation-guided matching approach.

\subsection{Deformation Guided Approaches}

Deformation guided approaches find a point-to-point map between two shapes by aligning them with a deformation in 3D. A typical example is the Iterative Closest Point~\cite{Besl1992} registration algorithm and its non-rigid sequel NICP~\cite{Amberg2007}.

Early deep deformation-guided strategies,~\eg~\cite{Groueix2018atlas, groueix20183d, deprelle2019learning}, learn to align point clouds to a common structure. 

Groueix~\etal~\cite{groueix2019unsupervised} propose an unsupervised network for point cloud matching by template-free cycle consistent deformations.
Transfomesh~\cite{trappolini2021shape} leverages the power of the attention mechanism~\cite{Vaswani2017} for the task of point cloud registration. Sundararaman~\etal~\cite{sundararaman2022implicit} propose a supervised approach that learns a volumetric flow-field deformation between signed distance representations of the input shapes. Marin~\etal~\cite{marin2024nicp} propose Neural ICP (NICP), a variant of ICP~\cite{Besl1992} for non-rigid point cloud registration to a template. Non-learning based ARC-Flow~\cite{hartshorne2025arc} proposes to align shapes using a topology preserving diffeomorphic vector field; this assumption fails when meshes differ in topology.

Other approaches, \eg~\cite{zeng2021corrnet3d, lang2021dpc, eisenberger2021neuromorph, Merrouche_2023_BMVC}, associate shapes in a discriminative feature space guided by their 3D alignment. Unsupervised point cloud matching strategies CorrNet3D~\cite{zeng2021corrnet3d} and DPC~\cite{lang2021dpc} link feature-similarity based associations with a point cloud alignment in 3D. While CorrNet3D leverages the reconstruction of the two point clouds, DPC uses their self- and cross-reconstruction. Point cloud approaches, while robust to topological artefacts, lead to lower performances compared to mesh counterparts. 

NeuroMorph~\cite{eisenberger2021neuromorph} proposes a similar strategy for unsupervised mesh matching by guiding feature-based associations with a 3D interpolation. The latter is constrained using the ARAP loss~\cite{sorkine2007rigid}. Merrouche~\etal~\cite{Merrouche_2023_BMVC} extend this idea using a hierarchy of surface patches to find multi-scale associations guided by multi-scale ARAP deformations. Both approaches use the mesh connectivity to define rigidity constraints, which may contradict the ground truth correspondences when the input meshes contain topological artefacts. 

Merrouche~\etal~\cite{Merrouche_2023_BMVC} tackle this with a multi-scale feature space learned on a dataset of shapes with correct topology. While the latter allows robustness to meshes with topological artefacts, this approach does not address the problem of aligning them in 3D. We show that while still competitive \wrt Merrouche~\etal~\cite{Merrouche_2023_BMVC}, our approach achieves better 3D alignments, all while not relying on any data-driven prior.

\subsection{Functional Maps}

Put forward by Ovsjiankov~\etal~\cite{ovsjanikov2012functional}, Functional Maps (FM) cast shape matching as a search for a map between the coefficients of real valued descriptors in the eigenfunctions of the Laplace Beltrami Operator (LBO)~\cite{Meyer2003, Pinkall1993} of the shapes. The FM based framework was extended to the data-driven regime by~Litany~\etal~\cite{litany2017deep} and to the unsupervised one~\cite{halimi2019unsupervised, ginzburg2020cyclic, roufosse2019unsupervised, sharma2020weakly}. Building on these seminal works, a rich line of research has emerged of both axiomatic.~\eg~\cite{ren2018continuous, melzi2019zoomout, eisenberger2020smooth, ren2021discrete} and data-driven strategies.~\eg~\cite{donati2020deep, li2022attentivefmaps, cao2023self, abdelreheem2023zero, sun2023spatially, cao2024spectral, cao2023unsupervised, magnet2024memory, Cao2022, zhuravlev2025denoising, eisenberger2020deep, efroni2022spectral, bastianxie2024hybrid}.

Cao~\etal~\cite{Cao2022} use unsupervised FMs to learn cycle consistent correspondences in a shape collection by learning point-to-point maps to a virtual template shape. Sun~\etal~\cite{sun2023spatially} propose a similar strategy using the characterisation of cycle consistent functional maps in the spectral domain. 

AttentiveFMaps~\cite{li2022attentivefmaps} uses a spectral attention framework to combine multiple resolution FMs to handle non-isometric shapes. DUO-FM~\cite{donati2022deep} uses complex FMs to estimate orientation preserving maps. Zhuravlev~\etal~\cite{zhuravlev2025denoising} infer FMs using denoising diffusion models.

Magnet~\etal~\cite{magnet2024memory} propose a differentiable version of the ZoomOut refinement algorithm~\cite{melzi2019zoomout} which is added as a layer in the deep FM pipeline. Cao~\etal~\cite{cao2023self} learns correspondences that generalise to both meshes and point clouds. It solves for the FM using deep features on the meshes and links it to a feature-similarity based point-to-point map on the point clouds. In addition to jointly training for a FM and for a feature-similarity based point-to-point map, ULRSSM~\cite{cao2023unsupervised} proposes to conduct test-time optimisation on shape pairs, leading to more robust correspondences.

Other approaches combine the alignment in the spectral shapes' space with an alignment in 3D. Eisenberger~\etal~ propose DeepShells~\cite{eisenberger2020deep} a deep learning version of the axiomatic approach SmoothShells~\cite{eisenberger2020smooth}. In both strategies, a FM is optimised with an ARAP alignment in 3D between the spectral reconstructions of the shapes. STS~\cite{efroni2022spectral} proposes a student-teacher mechanism where either DPC~\cite{lang2021dpc} or CorrNet3D~\cite{zeng2021corrnet3d} is used as a student network. Point cloud features extracted by the latter are used as functions to match in a teacher FM. SmS~\cite{cao2024spectral} jointly optimises for a FM and for an interpolation between the two shapes. It extracts deep features used both in the FM pipeline and to extract a point-to-point map linked to an interpolation search; similarly to deformation-guided Neuromorph~\cite{eisenberger2021neuromorph}.

LBO based FMs are sensitive to topological artefacts as the truncated LBO eigenfunctions are only stable under near-isometries.
Bastian~\etal~\cite{bastianxie2024hybrid} augment the LBO basis with extrinsic eigenfunctions from the elastic thin-shell Hessian, to gain more robustness. Our method outperforms Bastian~\etal~\cite{bastianxie2024hybrid} as shown in the experiments.

\section{Method}
\begin{figure*}[ht]
    \centering
    \includegraphics[width=1.0\textwidth]{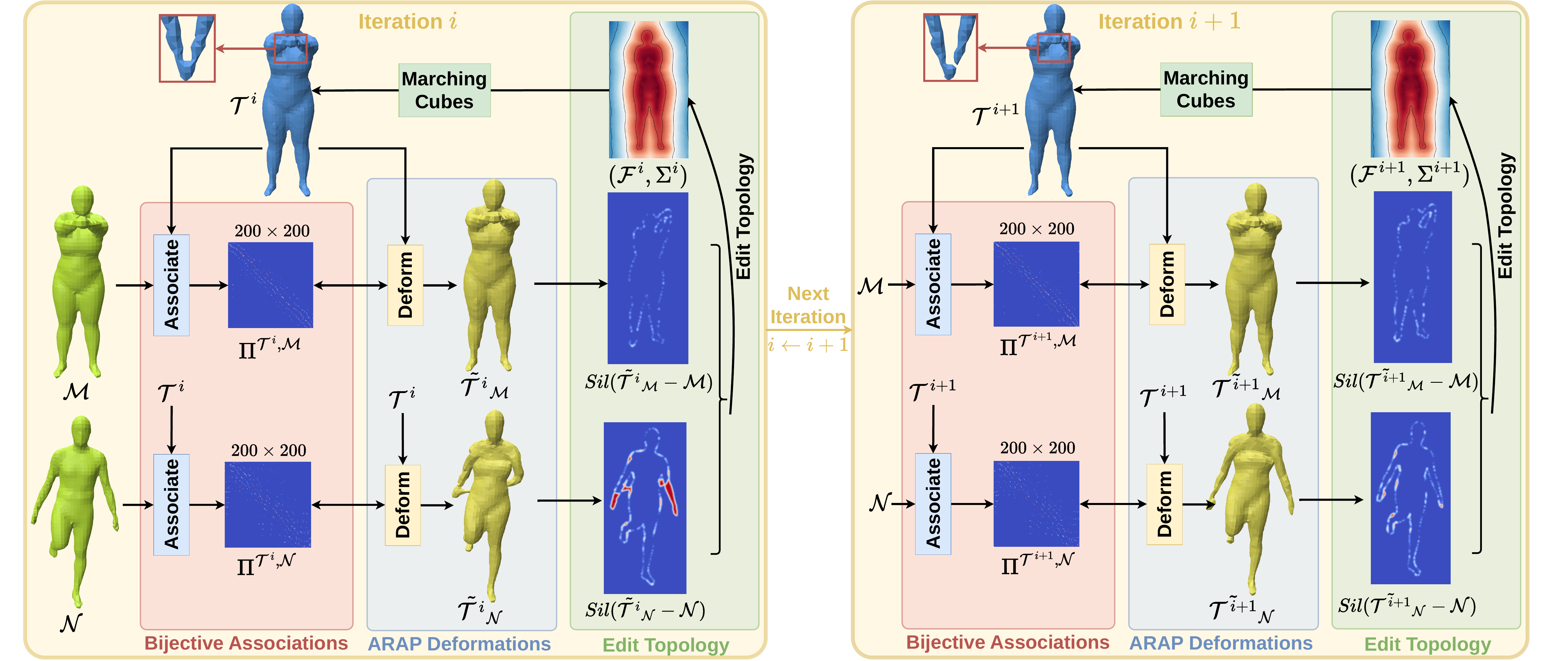}
    \caption{Given $\M$ and $\N$ with topological artefacts (glued hands in $\M$), we optimise for a template $\T$ and its alignments $\tilde{\T}_{\M}$ and $\tilde{\T}_{\N}$ with $\M$, $\N$. With $\T$ initialised as $\M$ or $\N$, we alternate between: 1. Finding associations $\Pi^{\T, \M}$ , $\Pi^{\T, \N}$ between $\T$ and $\M$, $\N$ as small matrices under a bijectivity constraint (red block), 2. Finding the best ARAP alignments $\tilde{\T}_{\M}$, $\tilde{\T}_{\N}$ between $\T$ and $\M$, $\N$ under these associations (blue block), 3. Editing $\T$'s topology when it impedes such alignments by updating its SDF representation $(\mathcal{F}, \Sigma)$ based on silhouettes (green block). Compare $\T^i$ and $\T^{i+1}$ and then $\tilde{\T^i}_{\N}$ and $\tilde{\T^{i+1}}_{\N}$.}
    \label{fig:method_overview}
\end{figure*}
Let $\M$ and $\N$ be 2D manifolds embedded in 3D space that might contain topological artefacts, discretised as 3D triangle meshes $\M = (V_{\M}, E_{\M})$ and $\N = (V_{\N}, E_{\N})$ respectively. Our goal is to find dense correspondences between $\M$ and $\N$. 

Our strategy finds a template mesh $\T = (V_{\T}, E_{\T})$ such that we can deform $\T$ with an \emph{as-rigid-as-possible deformation model under bijective associations} to get $\tilde{\T}_{\M} = (V_{\tilde{\T}_{\M}}, E_{\T})$ and $\tilde{\T}_{\N} = (V_{\tilde{\T}_{\N}}, E_{\T})$ such that $l(\tilde{\T}_{\M}, \M) \approx 0$ and $l(\tilde{\T}_{\N}, \N) \approx 0$, where $l$ is some measure of the alignment of two surfaces.

Vertex-to-vertex correspondences between $\M$ and $\N$,~\ie~$\phi : V_{\M} \to V_{\N}$, can be derived through the bijective map $C$ induced by $\T$ that maps the $i$-th vertex of $\tilde{\T}_{\M}$ to the $i$-th vertex of $\tilde{\T}_{\N}$, as follows: $\phi(x) = \arg\min_{w \in V_{\N}} \left\| w - C\left( \arg\min_{v \in V_{\tilde{\T}_{\M}}} \| x - v \|_2 \right) \right\|_2$. The same applies to get the correspondences from $\N$ to $\M$.

To this end we propose a topology-adaptive mesh deformation model~\ref{sec:topo_ada_def}, that is, a deformation model able to edit $\T$'s topology to achieve $l(\tilde{\T}_{\M}, \M) \approx 0$ and $l(\tilde{\T}_{\N}, \N) \approx 0$. Using this deformation model, we derive a shape matching approach~\ref{sec:opt_matching}. Fig.~\ref{fig:method_overview} shows an overview of our approach.

\subsection{Topology-Adaptive Association-Aware ARAP Deformation Model}
\label{sec:topo_ada_def}

Given two 3D triangle meshes $\T$ and $\M$ with topological artefacts, a deformation of $\T$ that achieves $l(\tilde{\T}_{\M}, \M)=0$ might not exist; one can think of cases where a split is necessary to align two surfaces. 

To address this, we jointly optimise for both $\tilde{\T}_{\M}$ and  $\T$, iterating between searching for $\tilde{\T}_{\M}$ that aligns the best with $\M$ and editing the topology of $\T$ where this alignment fails. 

Besides achieving $l(\tilde{\T}_{\M}, \M)=0$, we also seek that $\tilde{\T}_{\M}$ is an as-rigid-as-possible deformation of $\T$ and that it is induced by bijective associations between $\T$ and $\M$.

 To do so, we combine an ARAP-like mesh deformation model with a neural signed distance representation.
 On the one hand, mesh deformation modelling,~\eg~\cite{cagniart2010free}, allows to efficiently search for as-rigid-as-possible deformations by leveraging the mesh connectivity. We use a patch-wise rigid mesh deformation model. The same patches can be used for scale separation to define associations between $\M$ and $\T$ as a small matrix that conditions the deformation. 
 
 Neural signed distance representations on the other hand, allow for seamless changes in shape topology, when mesh representations rely on heavy algorithms~\eg~\cite{zaharescu2010topology}. Therefore, we adjoin to $\T$, a neural field representation where its topology can be edited when it impedes the alignment between $\tilde{\T}_{\M}$ and $\M$; $\T$ with edited topology can be recovered with an iso-surface extraction algorithm~\eg~\cite{mcubes}. 
 
 Sec~\ref{sec:neural_field_fitting} explains how we obtain the neural field representation; Sec~\ref{sec:assos_deform} introduces the association-aware ARAP mesh deformation model and Sec~\ref{sec:topo_update} describes how we combine both to perform topology editing steps.

\subsubsection{Neural Field Fitting}
\label{sec:neural_field_fitting}

Given $\T$, we fit a neural field whose zero level set corresponds to it. We represent a neural field with geometry aligned features that couple a feature volume $\mathcal{F}$ with a multi-layer perceptron (MLP) $S_{\Sigma}$. To get the SDF value at query point $x \in \mathbb{R}^3$, we trilinearly interpolate $\mathcal{F}$ at $x$ and pass the resulting feature to $S_{\Sigma}$ yielding the desired value. 
 We fit $\mathcal{F}$ and $\Sigma$ to $\T$ by minimising: 

{
\footnotesize
\begin{align}
    \label{fitting_loss}
    l_{fitting} =& \lambda_1 l_{SDF}(\mathcal{F}, \Sigma) + \lambda_2 l_{eikonal}(\mathcal{F}, \Sigma)
    \\ +& \lambda_3 l_{smoothing}(\mathcal{F}) + \lambda_4 l_{reg}(\mathcal{F}, \Sigma) \nonumber
\end{align}
}
where $\lambda_1,...\lambda_4 \in \mathbb{R}$ are weights for individual loss terms: $l_{SDF}$ encourages the neural field to represent the ground truth SDF defined by $\T$, $l_{eikonal}$ encourages the neural field to retain the SDF property, $l_{smoothing}$ promotes smooth feature volumes $\mathcal{F}$ and $l_{reg}$ is an $l_2$ regularisation on $\mathcal{F}$ and $\Sigma$.  These losses are detailed in Sec~\ref{sec:fitting_supp}.

Using this representation, $\T$'s topology can be altered by editing $(\mathcal{F}, \Sigma)$ and applying marching-cubes~\cite{mcubes}, all while ensuring that the resulting mesh is a discretisation of a 2D manifold as long as the editing maintains the SDF property.

\subsubsection{Association-Aware As-Rigid-As-Possible Deformation Model}
\label{sec:assos_deform}
 We use a patch-based mesh representation both to compactly define associations between $\M$ and $\T$ under a bijectivity constraint (Association Model Para.) and to define an as-rigid-as-possible deformation $\tilde{\T}_{\M}$ to align $\T$ with $\M$ while abiding by these associations (Deformation Model Para.).
 
$\T$ (resp. $\M$) is decomposed into $L$ non-overlapping surface patches $(P^{\T}_k)_{1 \leq k \leq L}$ (resp. $(P^{\M}_k)_{1 \leq k \leq L}$) with their centers $C^{\T} = (c^{\T}_k \in \mathbb{R}^3)_{1 \leq k \leq L}$ (resp. $C^{\M} = (c^{\M}_k \in \mathbb{R}^3)_{1 \leq k \leq L}$).

\noindent\textbf{Association Model}
Using the patch decomposition, we compactly represent associations between $\T$ and $\M$ using an $L \times L$ doubly stochastic (i.e., rows and columns are positive and sum to 1) matrix $\Pi^{\T,\M} \in \mathbb{R}^{L \times L}$. $\Pi^{\T,\M}$ associates the patches of $\T$ and $\M$. 

We do not directly optimise for the entries in $\Pi^{\T,\M}$, and rely on a parametrisation robust to both changes of the discretisation of $\T$ and $\M$ and changes of the patch decomposition (for a fixed $L$). This parametrisation acts as follows:
\begin{enumerate}
    \item Using a neural network $A_{\Theta}$ with parameters $\Theta$, we compute features for the patches of shapes $\T$ and $\M$ that we denote $F^{patch}_{\T} \in \mathbb{R}^{L \times d}$ and $F^{patch}_{\M} \in \mathbb{R}^{L \times d}$ respectively. $A_{\Theta}$ is a graph convolutional network using the patch decomposition for scale separation~\cite{Merrouche_2023_BMVC} taking as input vertex coordinates $V_{\T} \in \mathbb{R}^{|V_{\T}| \times 3}$, $V_{\M} \in \mathbb{R}^{|V_{\M}| \times 3}$ and vertex normals $N_{\T} \in \mathbb{R}^{|V_{\T}| \times 3}$, $N_{\M} \in \mathbb{R}^{|V_{\M}| \times 3}$ of $\T$ and $\M$ respectively.
    \item Compute inter-patch feature-similarity scores $S^{\T, \M} \in \mathbb{R}^{L \times L}$ using cosine similarity, as follows:
    $$s_{tm} := \frac{\langle{F}^{patch}_{\T,t},{F}^{patch}_{\M,m}\rangle_2}{	\|{F}^{patch}_{\T,t}\|_2\|{F}^{patch}_{\M,m}\|_2}$$
    \item Apply the sinkhorn operator~\cite{adams2011ranking, mena2018learning} to project $S^{\T, \M}$ onto the Birkhoff polytope to obtain an approximately doubly stochastic matrix $\Pi^{\T, \M}$. This avoids dealing with two different matrices (one row-normalized and one column-normalized) as in~\eg~\cite{Merrouche_2023_BMVC} and has been used for similar purposes~\eg~\cite{Cao2022, cao2023self}.
\end{enumerate}

The doubly stochastic inter-patch association matrix $\Pi^{\T, \M}$ has a direct geometric interpretation that we can use to define a corresponding 3D deformation. $\Pi^{\T, \M}$ can be considered as a matrix of barycentric weights mapping each of the patch centers $C^{\T}$ of $\T$ (resp. $C^{\M}$ of $\M$) to a barycentric combination of $C^{\M}$ (resp. $C^{\T}$) defined as $\Pi^{\T, \M} C^{\M}$ (resp. $\Pi^{\T, \M \top} C^{\T}$). 

$\Pi^{\T, \M} C^{\M}$ (resp. $\Pi^{\T, \M \top} C^{\T}$) are in the convex hull of $C^{\M}$ (resp. $C^{\T}$). 
Further, if $\Pi^{\T, \M}$ is a permutation matrix, $\Pi^{\T, \M} C^{\M}$ (resp. $\Pi^{\T, \M \top} C^{\T}$) are on the surface of $\M$ (resp. $\T$). When optimising for $\Theta$, we can use the characterisation of permutation matrices as the orthogonal matrices whose entries are all non-negative to push $\Pi^{\T, \M}$ to be close to a permutation, which corresponds to bijective patch associations. 

We use this geometric interpretation of $\Pi^{\T, \M}$ to link the inter-patch feature-similarity based association search to a deformation in 3D,~\ie~to link $\Pi^{\T, \M}$ and $\tilde{\T}_{\M}$. Denoting $C^{\tilde{\T}_{\M}}$ the patch centers of $\T$ after deformation, we define a deformation-guided bijective association search objective over $\Theta$ as follows:

{
\footnotesize
\begin{align}
    \label{eq:association_loss}
    l_{assos}(\T, \M) &= \gamma_1 l_{perm}(\Pi^{\T, \M}) \\
    &+ \gamma_2 l_{match}(\Pi^{\T, \M}, \tilde{\T}_{\M}, \M) \nonumber
\end{align}
}
where $\gamma_1,\gamma_2 \in \mathbb{R}$ are weights for individual loss terms. $l_{perm}$, $l_{match}$ are defined as follows:

{
\footnotesize
\begin{equation}
\label{eq:bijective_loss}
    l_{perm}(\Pi^{\T, \M}) =  \| \Pi^{\T, \M}\Pi^{\T, \M \top} - I \|^2_2,
\end{equation}
}

{
\footnotesize
\begin{equation}
\label{eq:mathing_loss}
    l_{match}(\Pi^{\T, \M}, \tilde{\T}_{\M}, \M)  =  \| \Pi^{\T, \M} C^{\M}  - C^{\tilde{\T}_{\M}} \|^2_2,
\end{equation}
}
where $I$ denotes the $L \times L$ identity matrix.

\noindent\textbf{Deformation Model}
Given the patch decomposition of $\T$, we use a patch-wise rigid deformation representation~\cite{cagniart2010free} that compactly represents a non-rigid deformation of $\T$ as a collection of patch-wise rigid deformations. Patches of $\T$ define rigid deformations as rotations $R = (r_k \in \mathbb{R}^{3 \times 3})_{1 \leq k \leq L}$ and translations $U = (u_k \in \mathbb{R}^3)_{1 \leq k \leq L}$ as well as a set of blending functions $(\alpha_k(v))_{1 \leq k \leq L}$ where $\alpha_k(v)$ depends on the euclidean distance of $v$ to $c_k$. The non-rigid deformation for $v \in P_i$ denoted $\tilde{v}$ is obtained by blending the rigid deformations as follows:

{
\footnotesize
\begin{equation}
    \label{eq:def_formula}
    \tilde{v} = \frac{1}{\sum_{P_j \in \mathcal{N}(P_i) \bigcup \{P_i\}}{\alpha_j(v)}} \sum_{P_j \in \mathcal{N}(P_i) \bigcup \{P_i\}}{\alpha_j(v)x_j(v)}
\end{equation}
}
where $x_j(v)=r_j (v - c_j) + u_j$ and $P_j \in \mathcal{N}(P_i)$ iff $v_i \in P_i$ and  $v_j \in P_j$ and $v_i$ and $v_j$ share an edge in $\T$. Using the patch neighborhoods, an as-rigid-as-possible criterion on $R$ and $U$ which can be combined with different deformation objectives, can be written as follows:

{
\footnotesize

\begin{equation}
    \label{rig_loss}
    l_{rig}(R, U) = \sum_{(P_k)_{1 \leq k \leq L}}{\sum_{P_j \in \mathcal{N}(P_k)}{\sum_{v \in P_k \bigcup P_j}{E^{kj}_v}}} \mbox{ with,}
\end{equation}

\begin{equation}
        E^{kj}_{v\in P_k \bigcup P_j} =  (\alpha_k(v) + \alpha_j(v)) \| x_k(v) - x_j(v) \|^2_2,
\end{equation}

}

Given $\T$, $\M$ and inter-patch associations $\Pi^{\T, \M}$, we use this deformation model to represent the deformation that aligns $\T$ with $\M$,~\ie~$\tilde{\T}_{\M}$. 

Instead of directly optimising for the patch-wise rotations and translations that we denote $R_{\T, \M}$ and $U_{\T, \M}$ respectively, we optimise for the parameters of a graph convolutional network that yields these rigid deformations~\cite{Merrouche_2023_BMVC}. This parametrisation is robust to both changes of the discretisation and of the patch segmentation (for a fixed $L$).

This neural network that we call $D_{\Psi}$ with parameters $\Psi$, consists of a convolutional graph neural network using the patches for scale separation followed by an MLP. It takes as input the vertex coordinates $V_{\T} \in \mathbb{R}^{|V_{\T}| \times 3}$, $V_{\M} \in \mathbb{R}^{|V_{\M}| \times 3}$ and vertex normals $N_{\T} \in \mathbb{R}^{|V_{\T}| \times 3}$, $N_{\M} \in \mathbb{R}^{|V_{\M}| \times 3}$ of $\T$ and $\M$ respectively, and the inter-patch associations $\Pi^{\T, \M} \in \mathbb{R}^{L \times L}$ and outputs patch-wise rotations $R_{\T, \M} \in \mathbb{R}^{L \times 3 \times 3}$ and translations $U_{\T, \M} \in \mathbb{R}^{L \times 3}$. Applying the latter on $\T$ using Eq.~\ref{eq:def_formula} leads to $\tilde{\T}_{\M}$. 

To assess how well $\tilde{\T}_{\M}$ aligns with $\M$,~\ie~$l$ so far,
we use the silhouette loss. Using a differentiable rasterizer~\cite{Laine2020diffrast}, we project $\M$ and $\tilde{\T}_{\M}$ into image space and measure the alignment of their silhouette images. The silhouette loss between $\M$ and $\tilde{\T}_{\M}$ captures the differences in shape topology between $\M$ and $\T$ by projecting the full surfaces. Other measures of the alignment of two meshes,~\eg~Chamfer distance, act on the point sets defined by $V_{\M}$ and $V_{\tilde{\T}_{\M}}$ and are agnostic to the way these points connect along the surface. 

We define an as-rigid-as-possible deformation search objective over $\Psi$ to align $\T$ with $\M$ under bijective associations as follows:

{
\footnotesize
\begin{align}
    \label{eq:def_loss}
    l_{def}(\T, \M, \Pi^{\T, \M}) &= \beta_1 l_{match}(\Pi^{\T, \M}, \tilde{\T}_{\M}, \M) \\
    &+ \beta_2 l_{sil}(\M, \tilde{\T}_{\M}) + \beta_3 l_{rig}(R_{\T, \M}, U_{\T, \M}) \nonumber
\end{align}
}
where $\beta_1,\beta_2, \beta_3 \in \mathbb{R}$ are weights for individual loss terms, $l_{match}$ is defined in Eq.~\ref{eq:mathing_loss}, $l_{rig}$ is defined in Eq.~\ref{rig_loss} and $l_{sil}$ is the silhouette loss written as follows:

{
\footnotesize

\begin{align}
\label{eq:sil_loss}
    {l}_{sil}(\M, \tilde{\T}_{\M}) &= \frac{1}{|\Omega|} \sum_{\omega \in \Omega} \frac{1}{HW} \sum_{p=1}^{H\times W} \Bigg(
     \\  & \left(  \left\| (\mathcal{G}_{k, \sigma}*P_{\omega}(\M))_{p} - (\mathcal{G}_{k, \sigma}*P_{\omega}(\tilde{\T}_{\M}))_{p} \right\|_1  \right)  \Bigg) \nonumber
\end{align}

}

where $\Omega$ is the set of viewpoints, $H \times W$ is the resolution of the silhouette images, $P_{\omega}(\mathcal{M}) \in \{0,1\}^{H \times W}$ is the binary silhouette projection of shape $\mathcal{M}$ from viewpoint $\omega$, $\left\|.\right\|_1$ denotes the $l_1$ loss and $*$ denotes the convolution operation. We apply Gaussian smoothing with kernel $\mathcal{G}_{k, \sigma}$ of size $k$ and standard deviation $\sigma$ on the silhouettes to reduce the sensitivity of the loss to individual pixel variations.

 The objectives $l_{assos}$ in Eq.~\ref{eq:association_loss} and $l_{def}$ in Eq.~\ref{eq:def_loss} can be optimised jointly to get an as-rigid-as-possible  alignment of $\T$ with $\M$ under a bijective association constraint. 
 
\noindent\textbf{Graph Neural Network Parametrisation} 
Using the GNN parametrisation of associations and deformations with only vertex coordinates and vertex normals as input, allows robustness to both changes of mesh discretisations and patch segmentations (for a fixed $L$), all while remaining inherently local as GNNs operate on a limited receptive field. As we intend to edit $\T$ while optimising its as-rigid-as-possible alignment with $\M$ under bijective associations, this becomes critical enabling stability during optimisation.

\subsubsection{Topology Updates}
\label{sec:topo_update}

In cases where $\T$ and $\M$ contain topological artefacts, finding an ARAP alignment under bijective associations with $l_{sil}(\tilde{\T}_{\M}, \M) \approx 0$ might not feasible. We propose to edit $\T$'s topology where this alignment is not possible. 

To do so, we do not edit the mesh representation of $\T$ directly. Rather, we edit $\T$'s topology by editing its neural filed representation,~\ie~$(\mathcal{F}, \Sigma)$ and then apply marching cubes~\cite{mcubes} to get $\T$ with the updated topology, while ensuring that the latter remains a discretisation of a 2D manifold. We call this a topology update step and achieve it by optimising $(\mathcal{F}, \Sigma)$ to minimise:

{
\footnotesize

\begin{align}
    \label{eq:top_def_loss}
    l_{topo}(\M, \tilde{\T}_{\M})  &= \alpha_1 {l}_{sil}(\M, \tilde{\T}_{\M})  + \alpha_2 l_{min}(\T, \T^0, \tau) \\ &+ \alpha_3 l_{eikonal}(\mathcal{F}, \Sigma) + \alpha_4 l_{smoothing}(\mathcal{F}) \nonumber
\end{align}
}
where $\alpha_1,...,\alpha_4 \in \mathbb{R}$ are weights for individual loss terms. $l_{eikonal}$, $l_{smoothing}$ are identical to those in Eq.~\ref{fitting_loss} and $l_{sil}$ is defined in Eq.~\ref{eq:sil_loss}. $\T^{0}$ denotes the initial $\T$ that did not undergo any topology update steps and $l_{min}$ is a regularizer that aims to prevent topology update steps from leading to a new $\T$ that deviates too much from the original one $\T^0$. 

Indeed, the goal of topology update steps is solely to perform topological changes to satisfy ${l}_{sil}$ and not to optimise for the deformation itself. We found the Chamfer distance with a tolerance parameter $\tau$ between $\T$ and $\T^0$ to be a simple yet effective regulariser, it is implemented as follows:

{
\footnotesize

\begin{align}
    \label{cham_reg_loss}
    l_{min}(\T, \T^0, \tau) &= \frac{1}{|V_{\T}|} \sum_{v \in V_{\T}} \left( \max\left(0, d(v, V_{\T^0}) - \tau \right) \right)^2 \\
    &+ \frac{1}{|V_{\T^0}|} \sum_{v^0 \in V_{\T^0}} \left( \max\left(0, d(v^0, V_{\T}) - \tau \right) \right)^2 \nonumber
\end{align}

\begin{equation}
    d(v, V_{\T^0}) = \min_{v^0 \in V_{\T^0}} \|v - v^0\|_2
\end{equation}
}

$l_{min}$ bounds the deviations of the vertices of $\T$ from the original vertices $\T^0$ while being agnostic to the way these two sets of points connect, thus, allowing for topology changes while keeping the geometries of $\T$ and $\T^0$ close.

Note that $l_{sil}$ and $l_{min}$ are defined on the vertices of $\T$ and not on $(\mathcal{F}, \Sigma)$. Still, one can define gradients of these two losses \wrt $(\mathcal{F}, \Sigma)$ using their gradients \wrt the vertices of $\T$ e.g.~\cite{remelli2020meshsdf}.

Starting from an initial $\T^0$, one can jointly optimise for $\tilde{\T}_{\M}$ and $\T$ by iterating over the following steps:

\begin{enumerate}
    \item Optimise $\Theta$ to minimise $l_{assos}$ in Eq.~\ref{eq:association_loss} to get bijective associations between $\T^i$ and $\M$.
    \item Optimise $\Psi$ to minimise $l_{def}$ in Eq.~\ref{eq:def_loss} to get the best as-rigid-as-possible alignment between $\T^i$ and $\M$,~\ie~$\tilde{\T^i}_{\M}$, under the bijective associations of step 1.
    \item Optimise $(\mathcal{F}, \Sigma)$ to minimise $l_{topo}$ in Eq.~\ref{eq:top_def_loss} to get $\T^{i+1}$.
\end{enumerate}

Topology updates in step 3 leverage the direction of the gradient of $l_{topo}$ \wrt the vertices of $\T^i$,~\ie~$V_{\T^i}$, to obtain $\T^{i+1}$. While this can lead to a $\T^{i+1}$ with a topology different from that of $\T^i$, the direction given by this gradient remains topology agnostic as it is unaware of $E_{\T^i}$. Thus step 3 is unaware of the impact of the topology changes it induces on steps 1 and 2. 

While being robust to changes in discretisation and inherently local, associations and deformations are, at least locally, affected by the topology changes in step 3. This is since we rely on the patch neighborhoods in the definition of the graph convolution's receptive filed in $A_{\Theta}$ and $D_{\Psi}$, in the deformation model's equation (Eq.~\ref{eq:def_formula}) and in the rigidity criterion (Eq.~\ref{rig_loss}). Thus, taking a step in the direction of the gradient of $l_{topo}$ \wrt $V_{\T^i}$ might worsen the alignment found in steps 1 and 2 by leading to unwanted topology changes.

To alleviate this issue, we propose a greedy hill climbing like topology optimisation strategy. After obtaining $\T^{i+1}$ by taking a step in the direction of the gradient of $l_{topo}$ \wrt $V_\T^i$; we perform steps 1 and 2 with $\T^{i+1}$. If $\T^{i+1}$ leads to a better alignment than the one obtained with $\T^{i}$,~\ie~$l_{sil}(\M, \tilde{\T^{i+1}}_{\M})<l_{sil}(\M, \tilde{\T^i}_{\M})$, than we perform step 3; otherwise we resume steps 1 and 2 with $\T^{i}$. 

In other words, a topology update step is effectively carried out, only if it allows improving the alignment with the target shape.

\subsection{Topology-Adaptive Deformation-Guided Mesh Matching}

\label{sec:opt_matching}
Using the association-aware as-rigid-as-possible topology-adaptive deformation model introduced in Sec.~\ref{sec:topo_ada_def}, we derive a matching strategy between meshes $\M$ and $\N$ that might contain topological artefacts.

We find a template mesh $\T = (V_{\T}, E_{\T})$ and association-aware as-rigid-as-possible deformations $\tilde{\T}_{\M} = (V_{\tilde{\T}_{\M}}, E_{\T})$ and $\tilde{\T}_{\N} = (V_{\tilde{\T}_{\N}}, E_{\T})$ that align $\T$ with $\M$ and $\N$ respectively. This is done by initialising $\T$ as either $\M$ or $\N$ and alternating between optimising for $\tilde{\T}_{\M}$ and $\tilde{\T}_{\N}$ and editing $\T$ where it impedes these alignments, as detailed below:

Initialising $\T^{0}$ as $\M$ and obtaining its neural field representation $(\mathcal{F}^0, S_{\Sigma^0})$ and setting $\tilde{\T^0}_{\M}=\T^0$ and $\tilde{\T^0}_{\N}=\T^0$, our strategy finds $\T=\T^n$, $\tilde{\T}_{\M}=\tilde{\T^n}_{\M}$ and $\tilde{\T}_{\N} = \tilde{\T^n}_{\N}$ by iterating over the following steps for \( i = 0, \dots, n \):
\begin{enumerate}
    \item Obtain bijective associations $\Pi^{\T^i, \M}$ and $\Pi^{\T^i, \N}$ between $\T^i$ and $\M$ and between $\T^i$ and $\N$ respectively, by optimising $\Theta^i$ to minimise $l_{assos}(\T^i, \M)+l_{assos}(\T^i, \N)$ leading to $\Theta^{i+1}$.
    \item Under the bijective associations found in step 1, find the best as-rigid-as-possible alignment between $\T^{i}$ and $\M$ and between $\T^{i}$ and $\N$,~\ie $\tilde{\T^{i}}_{\M}$ and $\tilde{\T^{i}}_{\N}$ respectively, by optimising $\Psi^{i}$ to minimise $l_{def}(\T^i, \M, \Pi^{\T^i, \M}) + l_{def}(\T^i, \N, \Pi^{\T^i, \N})$ leading to $\Psi^{i+1}$.
    \item Perform a topology update step on $(\mathcal{F}^i, \Sigma^i)$ to get $(\mathcal{F}^{i+1}, \Sigma^{i+1})$ and set $\T_{i+1} = \mcubes{(\mathcal{F}^{i+1}, S_{\Sigma^{i+1}})}$:
    \begin{enumerate}
        \item If \big($l_{sil}(\M, \tilde{\T^{i}}_{\M}) + l_{sil}(\N, \tilde{\T^{i}}_{\N})<l_{sil}(\M, \tilde{\T^{i-1}}_{\M}) + l_{sil}(\N, \tilde{\T^{i-1}}_{\N})$\big) or if $(i=0)$, than edit  $(\mathcal{F}^i, \Sigma^i)$ by taking a step in the direction given by the gradient of $l_{topo}(\N, \tilde{\T^{i}}_{\N})$ \wrt $V_{\T^i}$ leading to $(\mathcal{F}^{i+1}, \Sigma^{i+1})$.
        \item Otherwise, set $(\mathcal{F}^{i+1}, \Sigma^{i+1}) = (\mathcal{F}^{i-1}, \Sigma^{i-1})$, $\Theta^{i+1} = \Theta^{i}$ and $\Psi^{i+1} = \Psi^{i}$.
    \end{enumerate}
\end{enumerate}

We provide additional details on the optimization and implementation details in Sec.~\ref{sec:impelm_details}.

\section{Experiments}
This section shows the effectiveness of our approach in matching highly non-isometric shapes and shapes with topological artefacts. Ablations assessing the benefit of core components and additional results are in Sec.~\ref{sec:additional_results}.

\noindent\textbf{Datasets :} We evaluate our approach on widely used shape matching benchmarks of template-fitted re-meshed human meshes from FAUST~\cite{Bogo_faust, ren2018continuous} and SCAPE~\cite{anguelov2005scape, ren2018continuous} and animal meshes from SMAL~\cite{zuffi20173d, donati2022deep}. Shapes in these datasets share the same topology. We evaluate robustness to topological artefacts on a subset of the ExtFAUST test set corrupted with topological noise~\cite{basset2021neural, Merrouche_2023_BMVC}. Finally, we compare on per-frame multi-view reconstructions of humans in everyday clothing from 4DHumanOutfit~\cite{armando20234dhumanoutfit} plagued with acquisition noise, altering the geometry and topology. To evaluate the latter we use the SMPL~\cite{loper2015smpl} fittings from~\cite{Merrouche_2023_BMVC}.

\noindent\textbf{Evaluation Metrics :} Following prior works we use the Mean Geodesic Error (MGE). Denoting $\phi^{gt}$ the ground truth point-to-point map between $\M$ and $\N$, $\phi$ the estimated one and $d_{\N}$ the geodesic distance on shape $\N$, the MGE can be written as follows: 

{
\footnotesize
\begin{equation}
\label{mge_error}
	\operatorname{MGE}(\phi)
	:=
	\frac{1}{|V_{\M}|}
	\sum_{v \in V_{\M}}{d_{\N}(\phi(v), \phi^{gt}(v))},
\end{equation}
}

To evaluate the 3D alignments, we use the Chamfer distance (CD) on uniformly sampled meshes (10$k$ points) normalised so their bounding box's largest edge is $10$.

\noindent\textbf{Baselines :} we compare with a selection of approaches on standard benchmarks FAUST, SCAPE and SMAL reporting results from the literature. On 4DHumanOutfit and ExtFAUST containing topological artefacts, we compare with FM based methods ULRSSM~\cite{cao2023unsupervised}, SmS~\cite{cao2024spectral}, Bastian~\etal~\cite{bastianxie2024hybrid} (Hybrid ULRSSM) and deformation-guided method Merrouche~\etal~\cite{Merrouche_2023_BMVC} as they all target topological artefacts in their evaluation. We use pre-trained models on FAUST for ULRSSM, Hybrid ULRSSM and SmS. For Merrouche~\etal~\cite{Merrouche_2023_BMVC}, we use the model pre-trained on the uncorrupted ExtFAUST train set. While all four methods benefit from both training and test-time optimization, our approach uses optimization only. The code for non-learning based ARC-Flow~\cite{hartshorne2025arc} is not yet available; the authors report SOTA results on SMAL and clearly indicate that topological changes are failure cases of their approach.
\begin{table}[h]
\centering
\small
\begin{tabular}{@{}clccc@{}}
\toprule
\textbf{Trained} & \textbf{Method} & \textbf{FAUST} & \textbf{SCAPE} & \textbf{SMAL} \\
\midrule
\multicolumn{5}{c}{\rule{1cm}{0.5pt} Functional Maps Based \rule{1cm}{0.5pt} }  \\
\cmark & Deep Shells~\cite{eisenberger2020deep} & 1.7 & 2.5 & 29.3 \\
\cmark & Cao~\etal~\cite{Cao2022} & 1.5 & 2.0 & - \\
\cmark & DUO-FMNet~\cite{donati2022deep} & 2.5 & 2.6 & 6.7 \\
\cmark & AttentiveFMaps~\cite{li2022attentivefmaps} & 1.9 & 2.1 & 5.4 \\
\cmark & Zhuravlev~\etal~\cite{zhuravlev2025denoising} & 1.7 & 2.1 & 4.3 \\
\cmark & Sun~\etal~\cite{sun2023spatially} & 1.7 & 2.4 & 5.4 \\
\cmark & SmS~\cite{cao2024spectral} & \textbf{1.4} & \textbf{1.8} & \textbf{1.9} \\
\cmark & Magnet~\etal~\cite{magnet2024memory} & 1.9 & 2.4 & -\\
\cmark & ULRSSM~\cite{cao2023unsupervised} & 1.6 & 1.9 & 3.9 \\
\cmark & Bastian~\etal~\cite{bastianxie2024hybrid} & 1.5 & \textbf{1.8} & 3.3 \\
\xmark & BCICP~\cite{ren2018continuous} & 6.1 & 11.0 & - \\
\xmark & ZoomOut~\cite{melzi2019zoomout} & 6.1 & 7.5 & 38.4 \\
\xmark & DiscreteOp~\cite{ren2021discrete} & 5.6 & 13.1 & 38.1 \\
\xmark & SmoothShells~\cite{eisenberger2020smooth} & \underline{2.5} & \underline{4.2} & 30 \\
\multicolumn{5}{c}{\rule{1cm}{0.5pt} Deformation Guided \rule{1cm}{0.5pt} }  \\
\cmark & Neuromorph~\cite{eisenberger2021neuromorph} & 2.3 & 5.6 & 5.9 \\
\cmark & Merrouche~\etal~\cite{Merrouche_2023_BMVC} & - & - & 5.23 \\
\xmark & Ours & 5.84 & 8.06 &  \underline{9.45} \\
\addlinespace
\bottomrule
\end{tabular}
\caption{Mean geodesic error for baselines on the FAUST, SCAPE, and SMAL datasets. A checkmark indicates a learning-based method.  ``-'' indicates result not available in the literature. Best overall in bold. Best non-learning based underlined.}
\label{tab:comparison_bentchmark}
\end{table}

\begin{figure}[h]
    \centering
    \includegraphics[width=0.9\columnwidth]{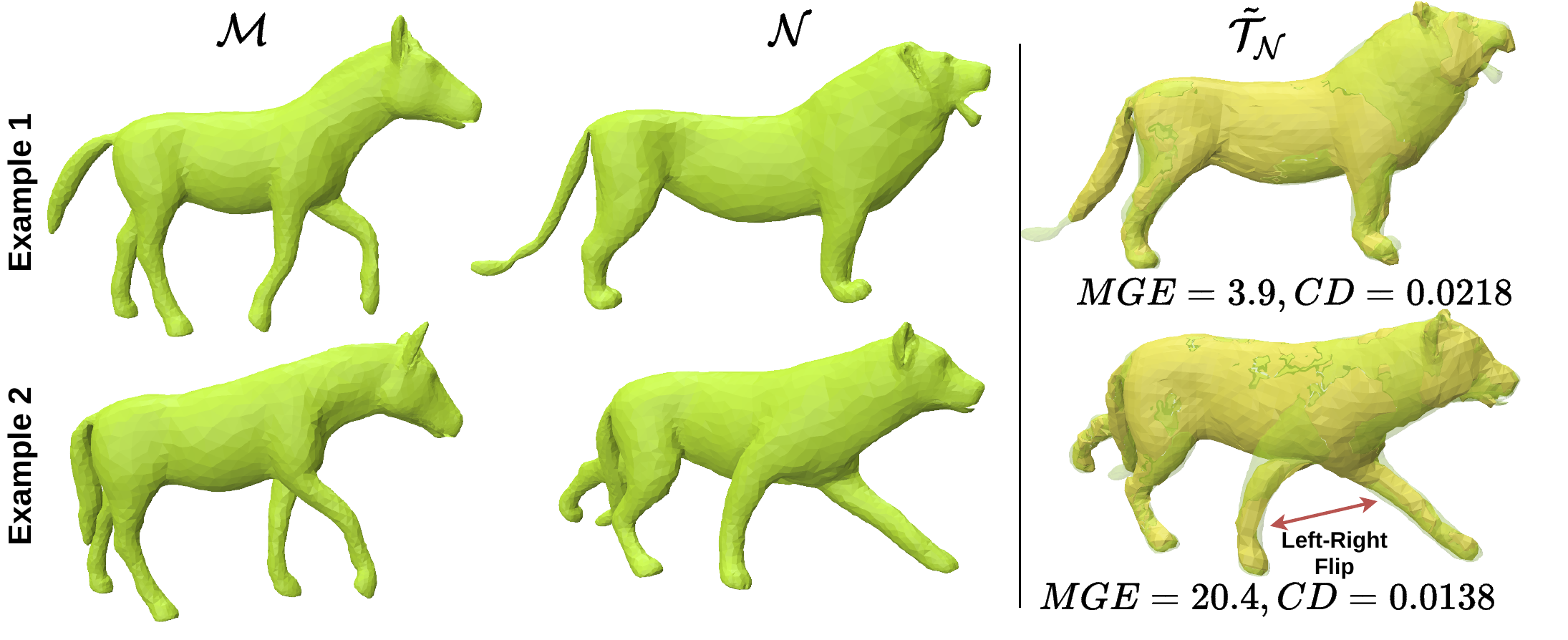}
    \caption{When a plausible alignment is possible, our approach can flip left and right which greatly worsens the geodesic error.}
    \label{fig:left_right_flips}
\end{figure}

\noindent\textbf{Benchmarks of Humans and Animals:} Tab.~\ref{tab:comparison_bentchmark} details the quantitative results on FAUST, SCAPE and SMAL. Our approach is the only non-learning based one that can handle non-isometric shapes from the SMAL dataset. Our approach can lead to left-right flips in cases when both coarse bijective associations and silhouette alignments are satisfiable. This can be seen in Fig~\ref{fig:left_right_flips}: compared to Example 1, Example 2  achieves a smaller Chamfer distance and a significantly larger geodesic error due to the left-right flip. This explains the difference in performance with learning based approaches across the benchmarks.
\begin{table}[h]
\centering
\footnotesize
\begin{tabular}{@{}cccccc@{}}
\toprule
\textbf{Trained} & \textbf{Method} & \multicolumn{2}{c}{\textbf{ExtFAUST}} & \multicolumn{2}{c}{\textbf{4DHOutfit}} \\
                 &                 & \textbf{MGE} & \textbf{CD}             & \textbf{MGE} & \textbf{CD} \\
\midrule
\multicolumn{5}{c}{\rule{1cm}{0.5pt} Functional Maps Based \rule{1cm}{0.5pt} }  \\

\cmark & ULRSSM~\cite{cao2023unsupervised} & 32.88 & - & 39.81 & - \\
\cmark & SmS~\cite{cao2024spectral} & 33.12 & - & 42.73 & - \\
\cmark & Bastian~\etal~\cite{bastianxie2024hybrid} & 8.76 & - & 30.11 & - \\
\multicolumn{5}{c}{\rule{1cm}{0.5pt} Deformation Guided \rule{1cm}{0.5pt} }  \\

\cmark & Merrouche~\etal~\cite{Merrouche_2023_BMVC} & \textbf{7.32} & 0.5795 & \textbf{6.15} & 0.2283 \\
\xmark & Ours & 8.74 & \textbf{0.3342} & 13.31 & \textbf{0.1597} \\
\addlinespace
\bottomrule
\end{tabular}
\caption{Mean geodesic error (MGE) and Chamfer Distance (CD) on the ExtFAUST and 4DHumanOutfit datasets. A checkmark indicates a learning-based method. Best in bold.}
\label{tab:comparison_ourdatasets}
\end{table}

\noindent\textbf{Humans with Topological Artefacts:} Tab.~\ref{tab:comparison_ourdatasets} details the quantitative results on ExtFAUST where shapes contain topological artefacts and multi-view human reconstructions in everyday clothing from 4DHumanOutfit plagued with acquisition noise. As expected, LBO based FM approaches ULRSSM and SmS fail in the presence of topological artefacts. Although on par with our approach on ExtFAUST, Hybrid ULRSSM's performances plummet on the most challenging shapes from 4DHumanOutfit. Merrouche~\etal~achieve the best performances in terms of geodesic errors while our approach achieves better alignments. 

\begin{figure}[h]
    \centering
    \includegraphics[width=0.9\columnwidth]{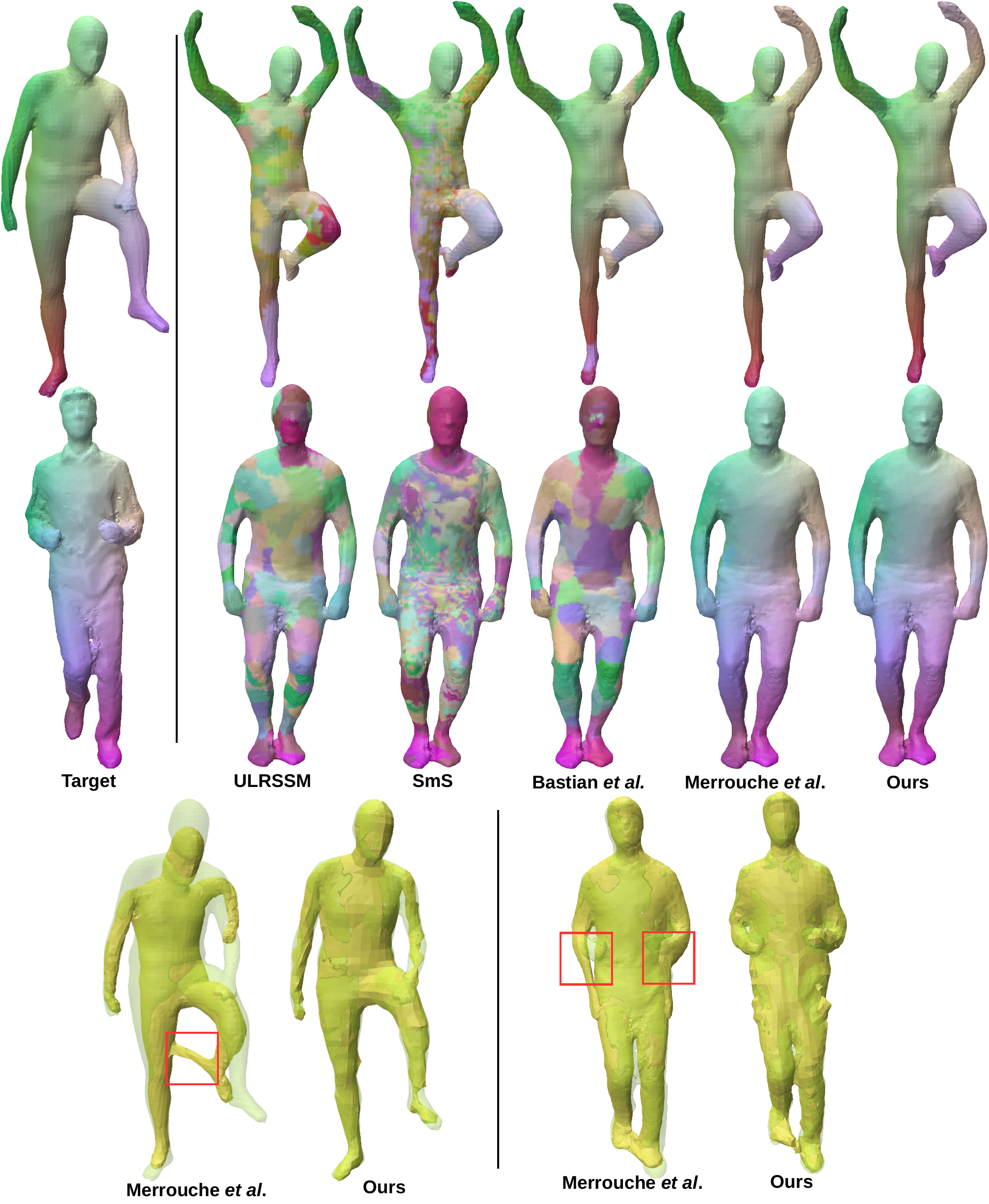}
    \caption{Correspondence results on ExtFAUST (top row) and 4DHumanOutfit (second row). The last row shows the 3D alignments found by the approach of Merrouche~\etal~\cite{Merrouche_2023_BMVC} and ours.}
    \label{fig:qual_results}
\end{figure}

This can be seen in Fig.~\ref{fig:qual_results}. The first and second row show an example from ExtFAUST and 4DHumanOutfit respectively. The colors are defined on the target shape and transferred to the source shape with the correspondences output by each method. Ours and Merrouche~\etal~ are the only approaches that handle both examples successfully. 

The last row shows the 3D alignments found by our method and Merrouche~\etal~ for both examples. Thanks to our topology adaptive deformation model, our strategy discards the topology of the source mesh when it impedes the matching objective (ankle detaches in first example, both hands detach in second one),  while the alignments found by Merrouche~\etal~suffer from the ARAP assumption when combined with topological artefacts (see red boxes). We highlight that contrary to Merrouche~\etal, our approach is purely optimisation based and does not benefit from any data-driven prior.

\section{Conclusions and Future Works}
While ubiquitous in real-world scenarios, topological artefacts pose a challenge to current mesh matching approaches. Leveraging a novel topology adaptive deformation model allowing edits in shape topology to align shape pairs under ARAP and bijective associations constraints, we tackle matching meshes that contain topological artefacts. 

We show that, while not relying on any data-driven prior, our approach applies to highly non-isometric shapes and shapes with topological artefacts including noisy per-frame multi-view reconstructions. Still, our approach could benefit from improvements.

To deal with left-right flips, the feature extractor with local receptive field in the association model can be augmented to capture more global shape properties. Further, both the association and the deformation networks can be pre-trained to benefit from priors learned on large shape collections.

While we employ the topology adaptive deformation model for mesh matching, its use could be extended to other setups,~\eg~multi-view reconstruction of moving shapes.

\section{Acknowledgements}
\paragraph{}
We thank Abdelmouttaleb Dakri for helpful discussions. This work was partially funded by the ANR project Human4D (ANR-19-CE23-0020).

{
    \small
    \bibliographystyle{plain}
    \bibliography{main}

\begin{thebibliography}{10}

\bibitem{abdelreheem2023zero}
Ahmed Abdelreheem, Abdelrahman Eldesokey, Maks Ovsjanikov, and Peter Wonka.
\newblock Zero-shot 3d shape correspondence.
\newblock In {\em SIGGRAPH Asia}, 2023.

\bibitem{Aberman2020}
Kfir Aberman, Peizh~Uo Li, Dani Lischinski, Olga Sorkine-Hornung, Daniel
  Cohen-Or, and Baoquan Chen.
\newblock Skeleton-aware networks for deep motion retargeting.
\newblock {\em ACM Transactions on Graphics (ToG)}, 2020.

\bibitem{adams2011ranking}
Ryan~Prescott Adams and Richard~S Zemel.
\newblock Ranking via sinkhorn propagation.
\newblock {\em arXiv preprint arXiv:1106.1925}, 2011.

\bibitem{Amberg2007}
Brian Amberg, Sami Romdhani, and Thomas Vetter.
\newblock Optimal step nonrigid icp algorithms for surface registration.
\newblock In {\em Conference on Computer Vision and Pattern Recognition}, 2007.

\bibitem{anguelov2005scape}
Dragomir Anguelov, Praveen Srinivasan, Daphne Koller, Sebastian Thrun, Jim
  Rodgers, and James Davis.
\newblock Scape: Shape completion and animation of people.
\newblock {\em ACM Transactions on Graphics (TOG)}, 2005.

\bibitem{armando20234dhumanoutfit}
Matthieu Armando, Laurence Boissieux, Edmond Boyer, Jean-S{\'e}bastien Franco,
  Martin Humenberger, Christophe Legras, Vincent Leroy, Mathieu Marsot, Julien
  Pansiot, Sergi Pujades, et~al.
\newblock 4dhumanoutfit: a multi-subject 4d dataset of human motion sequences
  in varying outfits exhibiting large displacements.
\newblock {\em Computer Vision and Image Understanding}, 2023.

\bibitem{Aubry2011}
Mathieu Aubry, Ulrich Schlickewei, and Daniel Cremers.
\newblock The wave kernel signature: A quantum mechanical approach to shape
  analysis.
\newblock In {\em International Conference on Computer Vision}, 2011.

\bibitem{basset2021neural}
Jean Basset, Adnane Boukhayma, Stefanie Wuhrer, Franck Multon, and Edmond
  Boyer.
\newblock Neural human deformation transfer.
\newblock In {\em International Conference on 3D Vision}, 2021.

\bibitem{bastianxie2024hybrid}
Lennart Bastian, Yizheng Xie, Nassir Navab, and Zorah L{\"a}hner.
\newblock Hybrid functional maps for crease-aware non-isometric shape matching.
\newblock In {\em Conference on Computer Vision and Pattern Recognition}, 2024.

\bibitem{Besl1992}
Paul~J. Besl and Neil~D. McKay.
\newblock A method for registration of 3-d shapes.
\newblock {\em IEEE Transactions on Pattern Analysis and Machine Intelligence},
  1992.

\bibitem{Biasotti2016}
S.~Biasotti, A.~Cerri, A.~Bronstein, and M.~Bronstein.
\newblock Recent trends, applications, and perspectives in 3d shape similarity
  assessment.
\newblock {\em Computer Graphics Forum}, 2016.

\bibitem{Bogo_faust}
Federica Bogo, Javier Romero, Matthew Loper, and Michael~J. Black.
\newblock {FAUST}: Dataset and evaluation for {3D} mesh registration.
\newblock In {\em Conference on Computer Vision and Pattern Recognition}, 2014.

\bibitem{Boscaini2016}
D.~Boscaini, J.~Masci, E.~Rodolà, M.~M. Bronstein, and D.~Cremers.
\newblock Anisotropic diffusion descriptors.
\newblock {\em Computer Graphics Forum}, 2016.

\bibitem{Boscaini20162}
Davide Boscaini, Jonathan Masci, Emanuele Rodolà, and Michael Bronstein.
\newblock Learning shape correspondence with anisotropic convolutional neural
  networks.
\newblock In {\em Advances in Neural Information Processing Systems}, 2016.

\bibitem{Bronstein2006}
Alexander~M. Bronstein, Michael~M. Bronstein, and Ron Kimmel.
\newblock Generalized multidimensional scaling: A framework for
  isometry-invariant partial matching.
\newblock {\em Proceedings of the National Academy of Sciences of the United
  States of America}, 2006.

\bibitem{Bronstein2010}
Alexander~M. Bronstein, Michael~M. Bronstein, Ron Kimmel, Mona Mahmoudi, and
  Guillermo Sapiro.
\newblock A gromov-hausdorff framework with diffusion geometry for
  topologically-robust non-rigid shape matching.
\newblock {\em International Journal of Computer Vision}, 2010.

\bibitem{cagniart2010free}
Cedric Cagniart, Edmond Boyer, and Slobodan Ilic.
\newblock Free-form mesh tracking: a patch-based approach.
\newblock In {\em Conference on Computer Vision and Pattern Recognition}, 2010.

\bibitem{Cao2022}
Dongliang Cao and Florian Bernard.
\newblock Unsupervised deep multi-shape matching.
\newblock In {\em Lecture Notes in Computer Science (including subseries
  Lecture Notes in Artificial Intelligence and Lecture Notes in
  Bioinformatics)}, 2022.

\bibitem{cao2023self}
Dongliang Cao and Florian Bernard.
\newblock Self-supervised learning for multimodal non-rigid 3d shape matching.
\newblock In {\em Conference on Computer Vision and Pattern Recognition}, 2023.

\bibitem{cao2024spectral}
Dongliang Cao, Marvin Eisenberger, Nafie El~Amrani, Daniel Cremers, and Florian
  Bernard.
\newblock Spectral meets spatial: Harmonising 3d shape matching and
  interpolation.
\newblock In {\em Conference on Computer Vision and Pattern Recognition}, 2024.

\bibitem{cao2023unsupervised}
Dongliang Cao, Paul Roetzer, and Florian Bernard.
\newblock Unsupervised learning of robust spectral shape matching.
\newblock {\em ACM Transactions on Graphics (ToG)}, 2023.

\bibitem{Coifman2005}
R.~R. Coifman, S.~Lafon, A.~B. Lee, M.~Maggioni, B.~Nadler, F.~Warner, and
  S.~W. Zucker.
\newblock Geometric diffusions as a tool for harmonic analysis and structure
  definition of data: Diffusion maps.
\newblock {\em Proceedings of the National Academy of Sciences of the United
  States of America}, 2005.

\bibitem{Deng2022}
Bailin Deng, Yuxin Yao, Roberto~M. Dyke, and Juyong Zhang.
\newblock A survey of non-rigid 3d registration.
\newblock {\em Computer Graphics Forum}, 2022.

\bibitem{deprelle2019learning}
Theo Deprelle, Thibault Groueix, Matthew Fisher, Vladimir Kim, Bryan Russell,
  and Mathieu Aubry.
\newblock Learning elementary structures for 3d shape generation and matching.
\newblock In {\em Advances in Neural Information Processing Systems}, 2019.

\bibitem{donati2022deep}
Nicolas Donati, Etienne Corman, and Maks Ovsjanikov.
\newblock Deep orientation-aware functional maps: Tackling symmetry issues in
  shape matching.
\newblock In {\em Conference on Computer Vision and Pattern Recognition}, 2022.

\bibitem{donati2020deep}
Nicolas Donati, Abhishek Sharma, and Maks Ovsjanikov.
\newblock Deep geometric functional maps: Robust feature learning for shape
  correspondence.
\newblock In {\em Conference on Computer Vision and Pattern Recognition}, 2020.

\bibitem{efroni2022spectral}
Omri Efroni, Dvir Ginzburg, and Dan Raviv.
\newblock Spectral teacher for a spatial student: Spectrum-aware real-time
  dense shape correspondence.
\newblock In {\em International Conference on 3D Vision}, 2022.

\bibitem{eisenberger2021neuromorph}
M.~Eisenberger, D.~Novotny, G.~Kerchenbaum, P.~Labatut, N.~Neverova,
  D.~Cremers, and A.~Vedaldi.
\newblock Neuromorph: Unsupervised shape interpolation and correspondence in
  one go.
\newblock In {\em Conference on Computer Vision and Pattern Recognition}, 2021.

\bibitem{eisenberger2020smooth}
Marvin Eisenberger, Zorah Lahner, and Daniel Cremers.
\newblock Smooth shells: Multi-scale shape registration with functional maps.
\newblock In {\em Conference on Computer Vision and Pattern Recognition}, 2020.

\bibitem{eisenberger2020deep}
Marvin Eisenberger, Aysim Toker, Laura Leal-Taix{\'e}, and Daniel Cremers.
\newblock Deep shells: Unsupervised shape correspondence with optimal
  transport.
\newblock In {\em Advances in Neural information processing systems}, 2020.

\bibitem{EladElbaz2003}
Asi~Elad Elbaz and Ron Kimmel.
\newblock On bending invariant signatures for surfaces.
\newblock {\em Transactions on Pattern Analysis and Machine Intelligence},
  2003.

\bibitem{ginzburg2020cyclic}
Dvir Ginzburg and Dan Raviv.
\newblock Cyclic functional mapping: Self-supervised correspondence between
  non-isometric deformable shapes.
\newblock In {\em European Conference on Computer Vision}, 2020.

\bibitem{groueix20183d}
Thibault Groueix, Matthew Fisher, Vladimir~G Kim, Bryan~C Russell, and Mathieu
  Aubry.
\newblock 3d-coded: 3d correspondences by deep deformation.
\newblock In {\em European Conference on Computer Vision}, 2018.

\bibitem{Groueix2018atlas}
Thibault Groueix, Matthew Fisher, Vladimir~G. Kim, Bryan~C. Russell, and
  Mathieu Aubry.
\newblock A papier-mache approach to learning 3d surface generation.
\newblock In {\em Conference on Computer Vision and Pattern Recognition}, 2018.

\bibitem{groueix2019unsupervised}
Thibault Groueix, Matthew Fisher, Vladimir~G Kim, Bryan~C Russell, and Mathieu
  Aubry.
\newblock Unsupervised cycle-consistent deformation for shape matching.
\newblock {\em Computer Graphics Forum}, 2019.

\bibitem{halimi2019unsupervised}
Oshri Halimi, Or~Litany, Emanuele Rodola, Alex~M Bronstein, and Ron Kimmel.
\newblock Unsupervised learning of dense shape correspondence.
\newblock In {\em Conference on Computer Vision and Pattern Recognition}, 2019.

\bibitem{hartshorne2025arc}
Adam Hartshorne, Allen Paul, Tony Shardlow, and Neill~DF Campbell.
\newblock Arc-flow: Articulated, resolution-agnostic, correspondence-free
  matching and interpolation of 3d shapes under flow fields.
\newblock In {\em International Conference on 3D Vision}, 2025.

\bibitem{kingma2014adam}
Diederik~P Kingma and Jimmy Ba.
\newblock Adam: A method for stochastic optimization.
\newblock {\em arXiv preprint arXiv:1412.6980}, 2014.

\bibitem{Laine2020diffrast}
Samuli Laine, Janne Hellsten, Tero Karras, Yeongho Seol, Jaakko Lehtinen, and
  Timo Aila.
\newblock Modular primitives for high-performance differentiable rendering.
\newblock {\em ACM Transactions on Graphics (ToG)}, 2020.

\bibitem{lang2021dpc}
Itai Lang, Dvir Ginzburg, Shai Avidan, and Dan Raviv.
\newblock Dpc: Unsupervised deep point correspondence via cross and self
  construction.
\newblock In {\em International Conference on 3D Vision}, 2021.

\bibitem{li2022attentivefmaps}
Lei Li, Nicolas Donati, and Maks Ovsjanikov.
\newblock Learning multi-resolution functional maps with spectral attention for
  robust shape matching.
\newblock In {\em Advances in Neural Information Processing Systems}, 2022.

\bibitem{Lim2020}
Jongin Lim, Hyung~Jin Chang, and Jin~Young Choi.
\newblock Pmnet: Learning of disentangled pose and movement for unsupervised
  motion retargeting.
\newblock In {\em British Machine Vision Conference}, 2020.

\bibitem{litany2017deep}
Or~Litany, Tal Remez, Emanuele Rodola, Alex Bronstein, and Michael Bronstein.
\newblock Deep functional maps: Structured prediction for dense shape
  correspondence.
\newblock In {\em International Conference on Computer Vision}, 2017.

\bibitem{Litman2011}
Roee Litman, Alexander~M. Bronstein, and Michael~M. Bronstein.
\newblock Diffusion-geometric maximally stable component detection in
  deformable shapes.
\newblock In {\em Computers and Graphics (Pergamon)}, 2011.

\bibitem{loper2015smpl}
Matthew Loper, Naureen Mahmood, Javier Romero, Gerard Pons-Moll, and Michael~J
  Black.
\newblock Smpl: A skinned multi-person linear model.
\newblock {\em ACM Transactions on Graphics (ToG)}, 2015.

\bibitem{mcubes}
William~E. Lorensen and Harvey~E. Cline.
\newblock Marching cubes: A high resolution 3d surface construction algorithm.
\newblock {\em ACM SIGGRAPH Computer Graphics}, 1987.

\bibitem{magnet2024memory}
Robin Magnet and Maks Ovsjanikov.
\newblock Memory-scalable and simplified functional map learning.
\newblock In {\em Conference on Computer Vision and Pattern Recognition}, 2024.

\bibitem{marin2024nicp}
Riccardo Marin, Enric Corona, and Gerard Pons-Moll.
\newblock Nicp: neural icp for 3d human registration at scale.
\newblock In {\em European Conference on Computer Vision}, 2024.

\bibitem{Masci2016}
Jonathan Masci, Davide Boscaini, Michael~M. Bronstein, and Pierre
  Vandergheynst.
\newblock Geodesic convolutional neural networks on riemannian manifolds.
\newblock In {\em International Conference on Computer Vision}, 2016.

\bibitem{melzi2019zoomout}
Simone Melzi, Jing Ren, Emanuele Rodolà, Abhishek Sharma, Peter Wonka, and
  Maks Ovsjanikov.
\newblock Zoomout: Spectral upsampling for efficient shape correspondence.
\newblock {\em ACM Transactions on Graphics (ToG)}, 2019.

\bibitem{mena2018learning}
Gonzalo Mena, David Belanger, Scott Linderman, and Jasper Snoek.
\newblock Learning latent permutations with gumbel-sinkhorn networks.
\newblock In {\em International Conference on Learning Representations}, 2018.

\bibitem{Merrouche_2023_BMVC}
Aymen Merrouche, Joao Pedro~Cova Regateiro, Stefanie Wuhrer, and Edmond Boyer.
\newblock Deformation-guided unsupervised non-rigid shape matching.
\newblock In {\em British Machine Vision Conference}, 2023.

\bibitem{Meyer2003}
Mark Meyer, Mathieu Desbrun, Peter Schr{\"o}der, and Alan~H Barr.
\newblock Discrete differential-geometry operators for triangulated
  2-manifolds.
\newblock In {\em Visualization and mathematics III}. Springer, 2003.

\bibitem{Monti2017}
Federico Monti, Davide Boscaini, Jonathan Masci, Emanuele Rodolà, Jan Svoboda,
  and Michael~M. Bronstein.
\newblock Geometric deep learning on graphs and manifolds using mixture model
  cnns.
\newblock In {\em Conference on Computer Vision and Pattern Recognition}, 2017.

\bibitem{ovsjanikov2012functional}
Maks Ovsjanikov, Mirela Ben-Chen, Justin Solomon, Adrian Butscher, and Leonidas
  Guibas.
\newblock Functional maps: a flexible representation of maps between shapes.
\newblock {\em ACM Transactions on Graphics (ToG)}, 2012.

\bibitem{Pinkall1993}
Ulrich Pinkall and Konrad Polthier.
\newblock Computing discrete minimal surfaces and their conjugates.
\newblock {\em Experimental Mathematics}, 1993.

\bibitem{Pishchulin2017}
Leonid Pishchulin, Stefanie Wuhrer, Thomas Helten, Christian Theobalt, and
  Bernt Schiele.
\newblock Building statistical shape spaces for 3d human modeling.
\newblock {\em Pattern Recognition}, 2017.

\bibitem{remelli2020meshsdf}
Edoardo Remelli, Artem Lukoianov, Stephan Richter, Benoit Guillard, Timur
  Bagautdinov, Pierre Baque, and Pascal Fua.
\newblock Meshsdf: Differentiable iso-surface extraction.
\newblock In {\em Advances in Neural Information Processing Systems}, 2020.

\bibitem{ren2021discrete}
Jing Ren, Simone Melzi, Peter Wonka, and Maks Ovsjanikov.
\newblock Discrete optimization for shape matching.
\newblock {\em Computer Graphics Forum}, 2021.

\bibitem{ren2018continuous}
Jing Ren, Adrien Poulenard, Peter Wonka, and Maks Ovsjanikov.
\newblock Continuous and orientation-preserving correspondences via functional
  maps.
\newblock {\em ACM Transactions on Graphics (ToG)}, 2018.

\bibitem{Rodol2014}
Emanuele Rodolà, Samuel~Rota Bulò, Thomas Windheuser, Matthias Vestner, and
  Daniel Cremers.
\newblock Dense non-rigid shape correspondence using random forests.
\newblock In {\em Conference on Computer Vision and Pattern Recognition}, 2014.

\bibitem{roufosse2019unsupervised}
Jean-Michel Roufosse, Abhishek Sharma, and Maks Ovsjanikov.
\newblock Unsupervised deep learning for structured shape matching.
\newblock In {\em International Conference on Computer Vision}, 2019.

\bibitem{Salti2014}
Samuele Salti, Federico Tombari, and Luigi~Di Stefano.
\newblock Shot: Unique signatures of histograms for surface and texture
  description.
\newblock {\em Computer Vision and Image Understanding}, 2014.

\bibitem{sharma2020weakly}
Abhishek Sharma and Maks Ovsjanikov.
\newblock Weakly supervised deep functional maps for shape matching.
\newblock In {\em Advances in Neural Information Processing Systems}, 2020.

\bibitem{sorkine2007rigid}
Olga Sorkine and Marc Alexa.
\newblock As-rigid-as-possible surface modeling.
\newblock In {\em Symposium on Geometry processing}, 2007.

\bibitem{Sun2009}
Jian Sun, Maks Ovsjanikov, and Leonidas Guibas.
\newblock A concise and provably informative multi-scale signature based on
  heat diffusion.
\newblock {\em Computer Graphics Forum}, 2009.

\bibitem{sun2023spatially}
Mingze Sun, Shiwei Mao, Puhua Jiang, Maks Ovsjanikov, and Ruqi Huang.
\newblock Spatially and spectrally consistent deep functional maps.
\newblock In {\em International Conference on Computer Vision}, 2023.

\bibitem{sundararaman2022implicit}
Ramana Sundararaman, Gautam Pai, and Maks Ovsjanikov.
\newblock Implicit field supervision for robust non-rigid shape matching.
\newblock In {\em European Conference on Computer Vision}, 2022.

\bibitem{trappolini2021shape}
Giovanni Trappolini, Luca Cosmo, Luca Moschella, Riccardo Marin, Simone Melzi,
  and Emanuele Rodol{\`a}.
\newblock Shape registration in the time of transformers.
\newblock In {\em Advances in Neural Information Processing Systems}, 2021.

\bibitem{vanKaick2011}
Oliver van Kaick, Hao Zhang, Ghassan Hamarneh, and Daniel Cohen-Or.
\newblock A survey on shape correspondence.
\newblock In {\em Eurographics Symposium on Geometry Processing}, 2011.

\bibitem{Vaswani2017}
Ashish Vaswani, Noam Shazeer, Niki Parmar, Jakob Uszkoreit, Llion Jones,
  Aidan~N. Gomez, Łukasz Kaiser, and Illia Polosukhin.
\newblock Attention is all you need.
\newblock In {\em Advances in Neural Information Processing Systems}, 2017.

\bibitem{verma2018feastnet}
Nitika Verma, Edmond Boyer, and Jakob Verbeek.
\newblock Feastnet: Feature-steered graph convolutions for 3d shape analysis.
\newblock In {\em Conference on Computer Vision and Pattern Recognition}, 2018.

\bibitem{Vestner2018}
Matthias Vestner, Zorah Lahner, Amit Boyarski, Or~Litany, Ron Slossberg, Tal
  Remez, Emanuele Rodola, Alex Bronstein, Michael Bronstein, Ron Kimmel, and
  Daniel Cremers.
\newblock Efficient deformable shape correspondence via kernel matching.
\newblock In {\em International Conference on 3D Vision}, 2018.

\bibitem{Vestner2017}
Matthias Vestner, Roee Litman, Emanuele Rodolà, Alex Bronstein, and Daniel
  Cremers.
\newblock Product manifold filter: Non-rigid shape correspondence via kernel
  density estimation in the product space.
\newblock In {\em Conference on Computer Vision and Pattern Recognition}, 2017.

\bibitem{zaharescu2010topology}
Andrei Zaharescu, Edmond Boyer, and Radu Horaud.
\newblock Topology-adaptive mesh deformation for surface evolution, morphing,
  and multiview reconstruction.
\newblock {\em IEEE Transactions on Pattern Analysis and Machine Intelligence},
  2010.

\bibitem{Zaharescu2009}
Andrei Zaharescu, Edmond Boyer, Kiran Varanasi, and Radu Horaud.
\newblock Surface feature detection and description with applications to mesh
  matching.
\newblock In {\em Conference on Computer Vision and Pattern Recognition}, 2009.

\bibitem{zeng2021corrnet3d}
Yiming Zeng, Yue Qian, Zhiyu Zhu, Junhui Hou, Hui Yuan, and Ying He.
\newblock Corrnet3d: Unsupervised end-to-end learning of dense correspondence
  for 3d point clouds.
\newblock In {\em Conference on Computer Vision and Pattern Recognition}, 2021.

\bibitem{zhou2019continuity}
Yi~Zhou, Connelly Barnes, Jingwan Lu, Jimei Yang, and Hao Li.
\newblock On the continuity of rotation representations in neural networks.
\newblock In {\em Conference on Computer Vision and Pattern Recognition}, 2019.

\bibitem{zhuravlev2025denoising}
Aleksei Zhuravlev, Zorah L{\"a}hner, and Vladislav Golyanik.
\newblock Denoising functional maps: Diffusion models for shape correspondence.
\newblock In {\em Conference on Computer Vision and Pattern Recognition}, 2025.

\bibitem{zuffi20173d}
Silvia Zuffi, Angjoo Kanazawa, David~W Jacobs, and Michael~J Black.
\newblock 3d menagerie: Modeling the 3d shape and pose of animals.
\newblock In {\em Conference on Computer Vision and Pattern Recognition}, 2017.

\end{thebibliography}
}

\clearpage
\setcounter{page}{1}
\maketitlesupplementary

This supplementary material presents additional quantitative and qualitative results including ablations showing the added benefit of the main components of our approach in Sec.~\ref{sec:additional_results} and implementation details in Sec.~\ref{sec:impelm_details}. Our code is publicly available at \href{https://gitlab.inria.fr/amerrouc/topology-adaptive-deformation-guided-matching}{https://gitlab.inria.fr/amerrouc/topology-adaptive-deformation-guided-matching}

\section{Additional Results}
\label{sec:additional_results}
\subsection{Ablation Studies}
We assess with ablations the added benefit of our approach's core components: the bijective associations assumption, the as-rigid-as-possible deformation assumption and the adaptive topology strategy. To ablate the bijective associations assumption, we optimise without $l_{perm}$ (Eq.~\ref{eq:bijective_loss}). For the ARAP assumption, we optimise without $l_{rig}$ (Eq.~\ref{rig_loss}). For the adaptive topology strategy, we optimise without the topology update steps (Sec.~\ref{sec:topo_update}). Table~\ref{tab:ablation_study} shows the results.

\begin{table}[h]
\centering
\small
\begin{tabular}{@{}ccccc@{}}
\toprule 
\makecell{Bijective\\Associations} & \makecell{ARAP\\Deformations} &  \makecell{Adaptive\\Topology} & \textbf{MGE}  & \textbf{CD}
\\ 
\midrule
\xmark     & \checkmark & \checkmark & {10.89} & {0.9545}\\
\checkmark & \xmark & \checkmark & {12.19} & \textbf{0.2129}\\
\checkmark & \checkmark & \xmark & 9.40 & 0.3419\\
\checkmark & \checkmark & \checkmark & \textbf{8.74} & {0.3342}\\
\addlinespace
\bottomrule
\end{tabular}
\caption{Ablations showing the impact of the bijective associations assumption, the as-rigid-as-possible deformation assumption and the topology update steps on the ExtFAUST dataset. Best in bold.}
\label{tab:ablation_study}
\end{table}

Our full optimisation achieves the best correspondences in terms of geodesic error, and the second best alignment quality in terms of Chamfer distance, after the one with ARAP ablated; as the latter allows vertices to freely move to align the deforming mesh with the target shapes leading to a lower Chamfer distance while drifting away from the ground truth correspondences.

\begin{figure}[h]
    \centering
    \includegraphics[width=1.0\columnwidth]{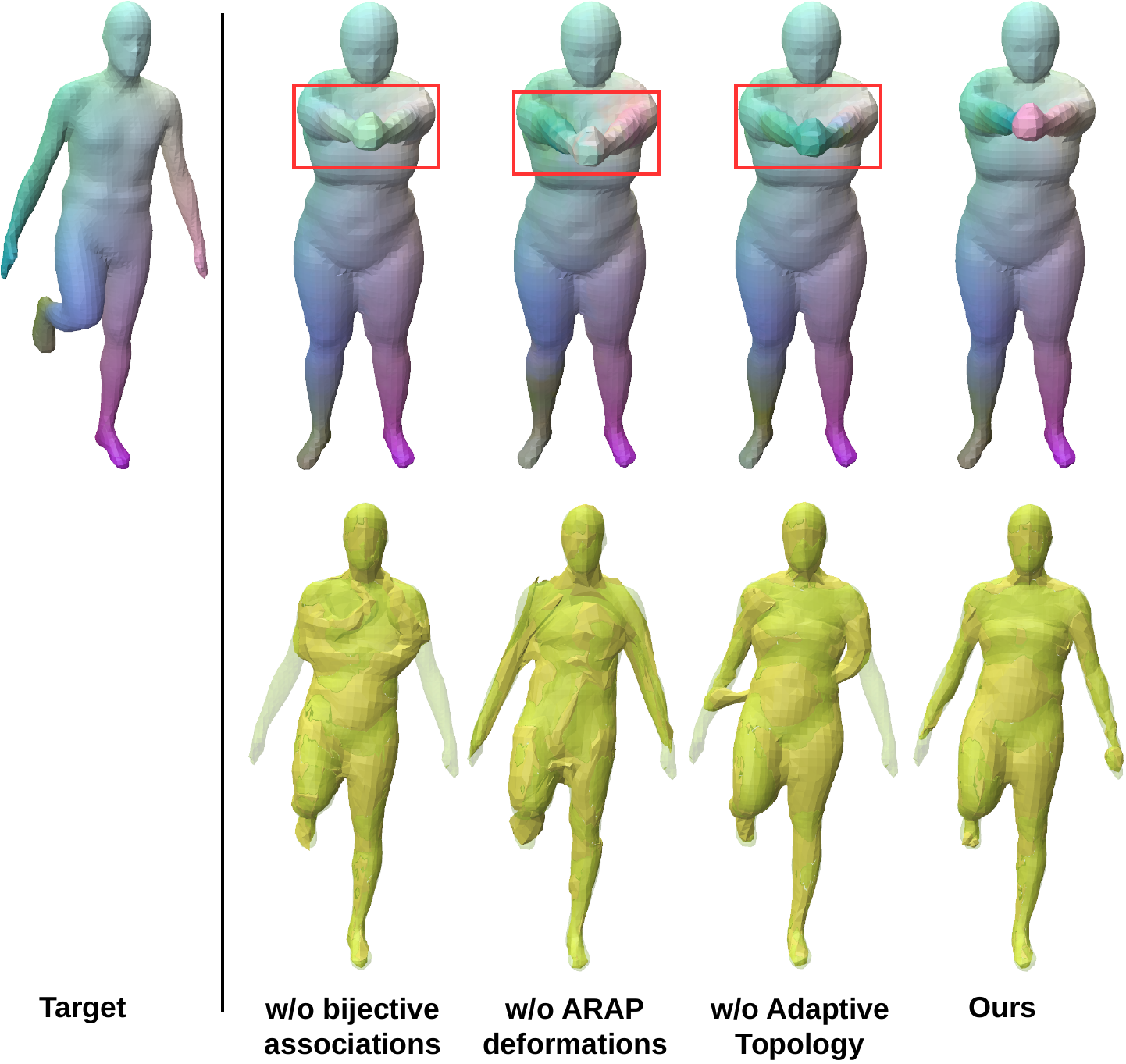}
    \caption{Qualitative ablation results on an example from ExtFAUST. The top row shows correspondences where colors are defined on the target mesh and transferred to the source mesh using the found alignments shown in the bottom row.}
    \label{fig:ablation_result}
\end{figure}

This can be seen in Fig.~\ref{fig:ablation_result}. Without the bijective associations constraint ensuring full coverage of the target shape, the optimisation can get stuck in local minima. Without the ARAP assumption, the deforming surface is distorted to align with the target, breaking the spatial continuity of the correspondences. Without an adaptive topology, an ARAP alignment under bijective associations is not feasible in some cases where the shapes contain topological artefacts.

We further show in Table~\ref{tab:ablation_study_losses} the effect of the periodic weighting scheme $w_K(i)$ detailed in Sec.~\ref{sec:opt_details} and the regulariser $l_{min}$ in Eq.~\ref{cham_reg_loss}.

\begin{table}[h]
\centering
\small
\begin{tabular}{@{}cccc@{}}
\toprule
\makecell{\textbf{$w_K(i)$}} &  \makecell{$l_{min}$} & \textbf{MGE}  & \textbf{CD}
\\ 
\midrule
\xmark     & \checkmark & 9.21 & {0.2360}\\
\checkmark & \xmark     & 9.14 & \textbf{0.2189} \\
\checkmark & \checkmark & \textbf{8.74} & {0.3342}\\
\addlinespace
\bottomrule
\end{tabular}
\caption{Ablations showing the impact of the periodic weighting scheme $w_K(i)$ (Sec.~\ref{sec:opt_details}) and the topology update steps' regulariser $l_{min}$ on the ExtFAUST dataset. Best in bold.}
\label{tab:ablation_study_losses}
\end{table}

Although leading to an increase in terms of Chamfer distance, optimising with both $w_K(i)$ and $l_{min}$ yields more accurate correspondences.

\subsection{Patch-Based Parametrisation}

\begin{figure}[h]
    \centering
    \includegraphics[width=1.0\columnwidth]{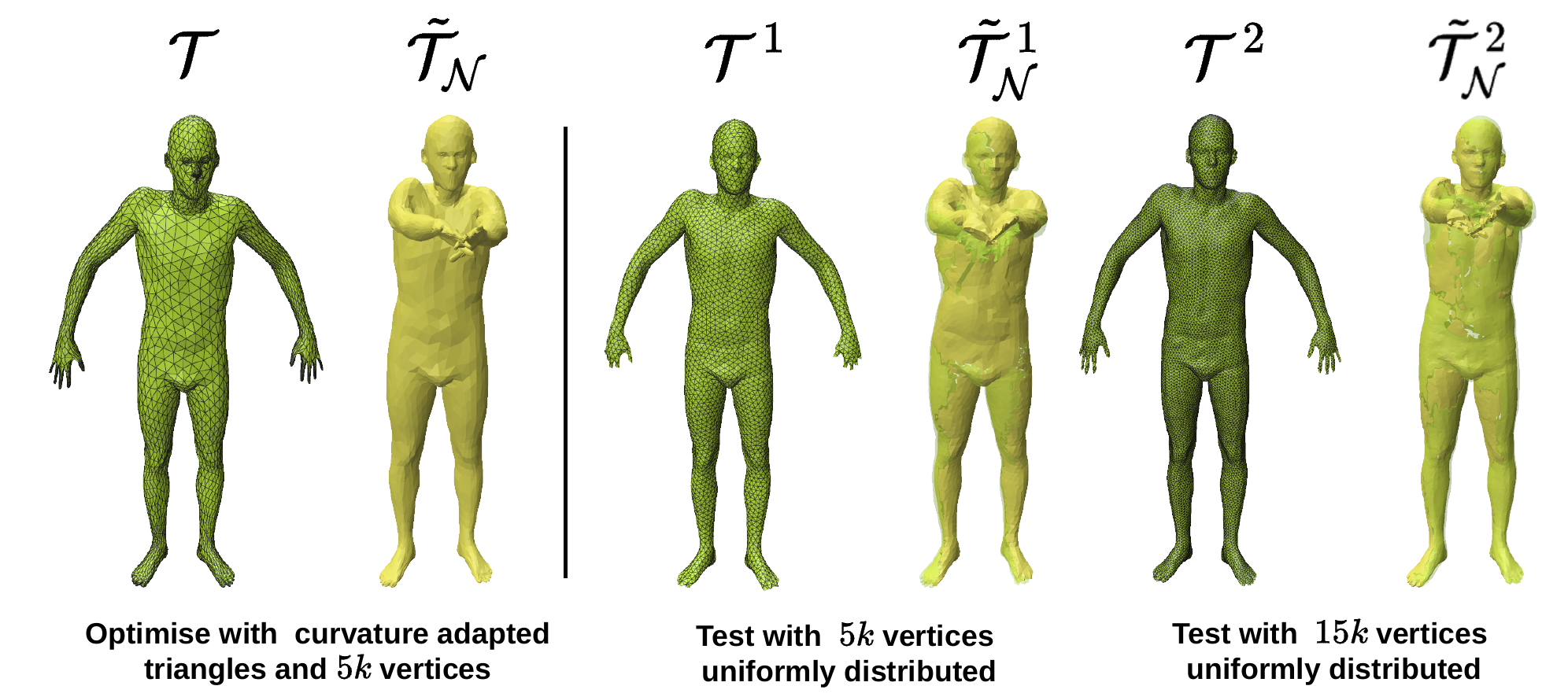}
    \caption{Robustness to discretisation of GNN parametrisation of associations and deformations. Associations and deformations are optimised on $\T$ and $\N$ and then tested without adaptation on $\T^1$, $\N$ and $\T^2$, $\N$.}
    \label{fig:gcnn_param_robus}
\end{figure}

We use graph convolution based neural networks both to parameterise the associations and the deformations, as it allows for some robustness to changes in mesh discretisation and patch decompositions (for a fixed $L$) which is essential to stabilise the optimisation. We show this on an extreme case in Fig.~\ref{fig:gcnn_param_robus}. We optimise for associations and deformations between two meshes  $\T$ and $\N$ sampled with curvature adapted triangles to contain $5k$ vertices. $\T$ and its alignment with $\N$, \ie $\tilde{\T}_\N$ can be seen on the left side of the figure. We directly test the same association and deformation networks on $\T^1$, $\N$ and $\T^2$, $\N$ without adaptation, where $\T^1$ and $\T^2$ are uniform resamplings of $\T$ with $5k$ and $15k$ vertices, respectively. $\T$, $\T^1$ and $\T^2$ have distinct patch decompositions. The resulting alignments $\tilde{\T^1}_{\N}$ and $\tilde{\T^2}_{\N}$, on the right side of the figure, show little differences to $\tilde{\T}_\N$ despite the significant change in discretisation.

Note that an ablation of this parametrisation is challenging: if associations and deformations are tied to the patch decomposition of a specific discretisation, they would need to be redefined after each topology update step.

\subsection{Qualitative Results on the Benchmarks}

\begin{figure}[h]
    \centering
    \includegraphics[width=1.0\columnwidth]{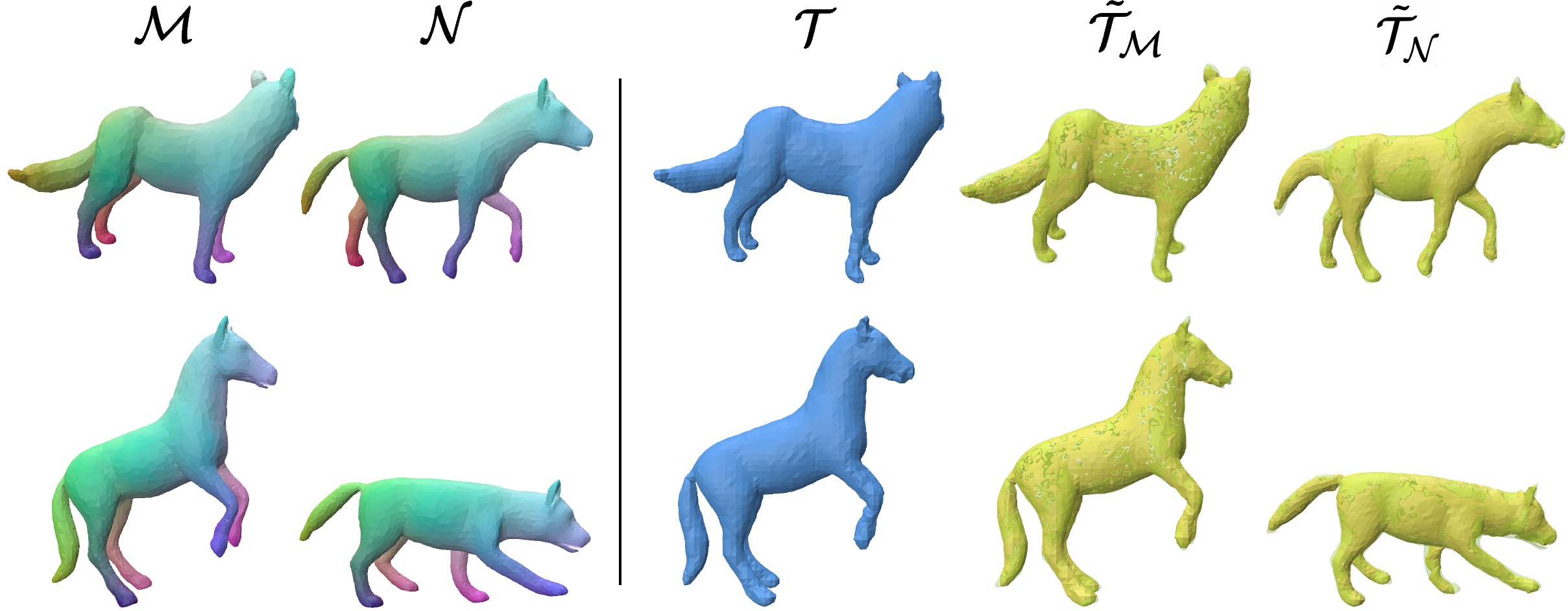}
    \caption{Correspondences and deformation results on examples from SMAL. Colors are defined on $\N$ and transferred to $\M$ using the found alignments $\tilde{\T}_{\M}$ and $\tilde{\T}_{\N}$. $\T$ is initialised as $\M$.}
    \label{fig:smal_examples}
\end{figure}

\begin{figure}[h]
    \centering
    \includegraphics[width=1.0\columnwidth]{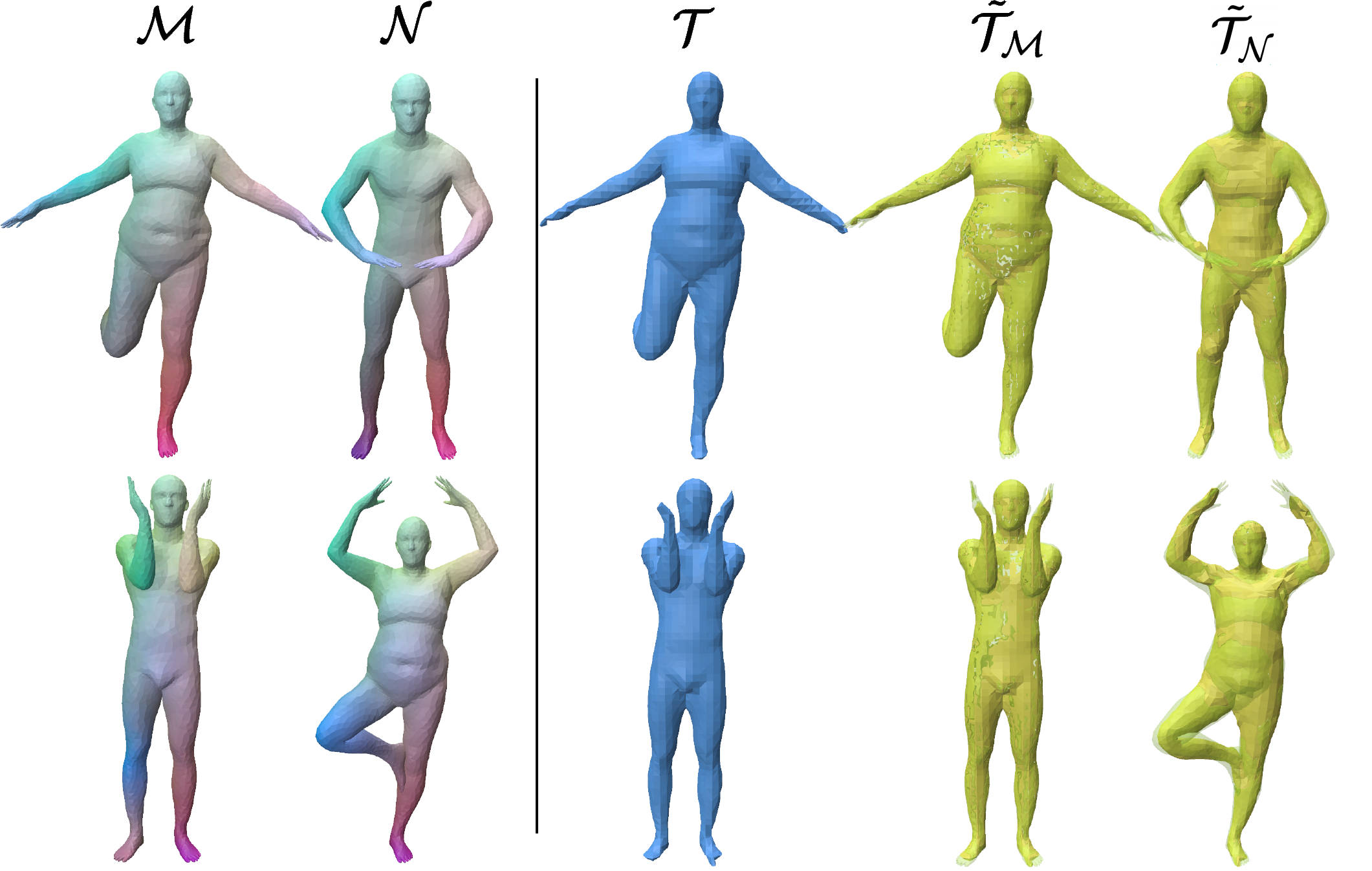}
    \caption{Correspondences and deformation results on examples from FAUST. Colors are defined on $\N$ and transferred to $\M$ using the found alignments $\tilde{\T}_{\M}$ and $\tilde{\T}_{\N}$. $\T$ is initialised as $\M$.}
    \label{fig:faust_examples}
\end{figure}

\begin{figure}[h]
    \centering
    \includegraphics[width=1.0\columnwidth]{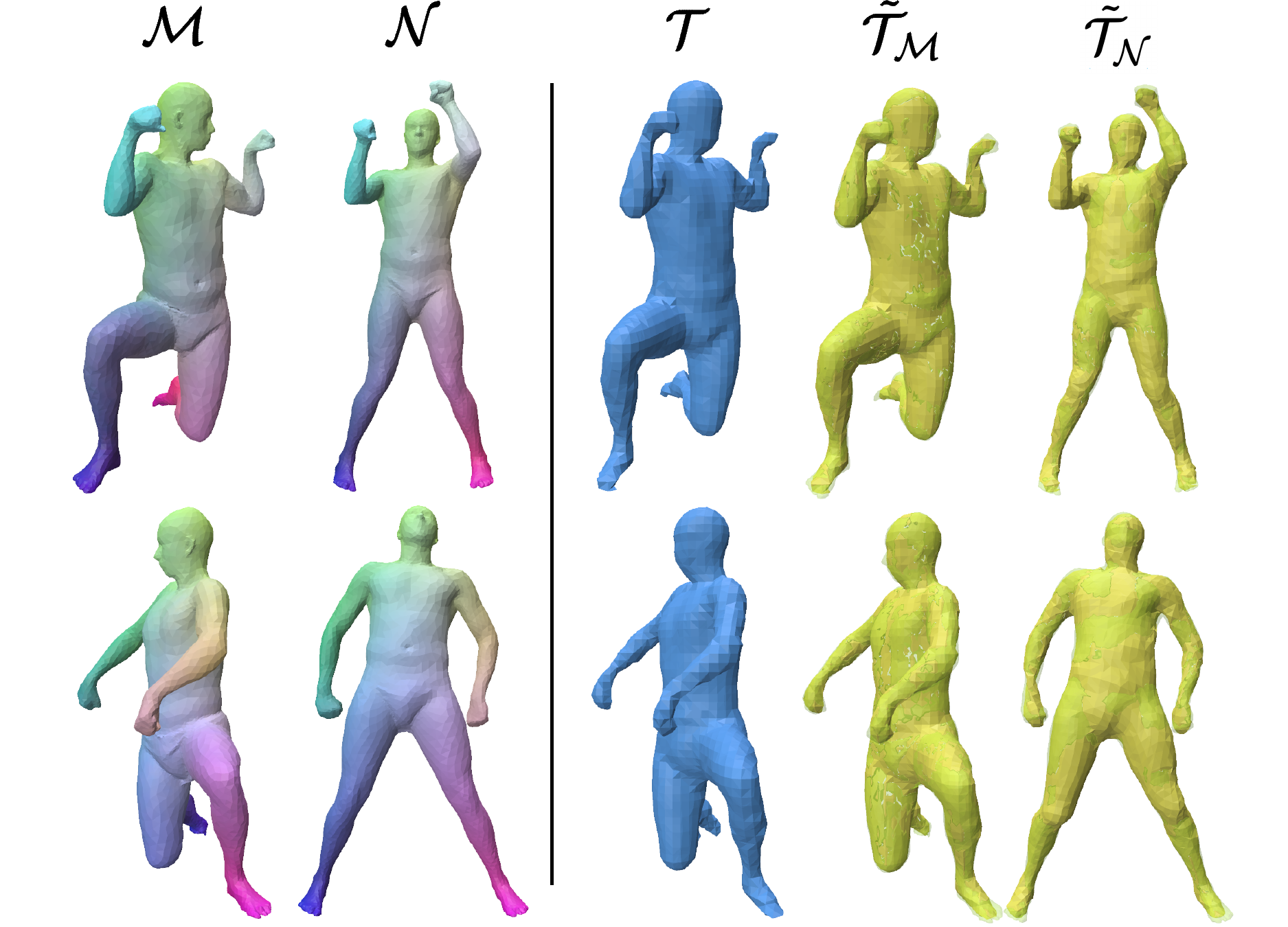}
    \caption{Correspondences and deformation results on examples from SCAPE. Colors are defined on $\N$ and transferred to $\M$ using the found alignments $\tilde{\T}_{\M}$ and $\tilde{\T}_{\N}$. $\T$ is initialised as $\M$.}
    \label{fig:scape_examples}
\end{figure}

We show qualitative correspondences and deformation results on the FAUST~\cite{Bogo_faust, ren2018continuous}, SCAPE~\cite{anguelov2005scape, ren2018continuous} and SMAL~\cite{zuffi20173d, donati2022deep}
 benchmarks. Fig.~\ref{fig:smal_examples} shows our results on highly non-isometric animals from SMAL, Fig.~\ref{fig:faust_examples} shows our results on humans with different identities and poses from FAUST and Fig.~\ref{fig:scape_examples} shows our results on human poses from SCAPE. In all figures, the colors are defined on $\N$ and transferred to $\M$ using the found alignments. The template $\T$ is initialised as $\M$. Note that, as all shapes in these datasets share the same topology, the template shows little differences to $\M$.

\subsection{Additional Qualitative Comparisons}

\begin{figure}[h]
    \centering
    \includegraphics[width=1.0\columnwidth]{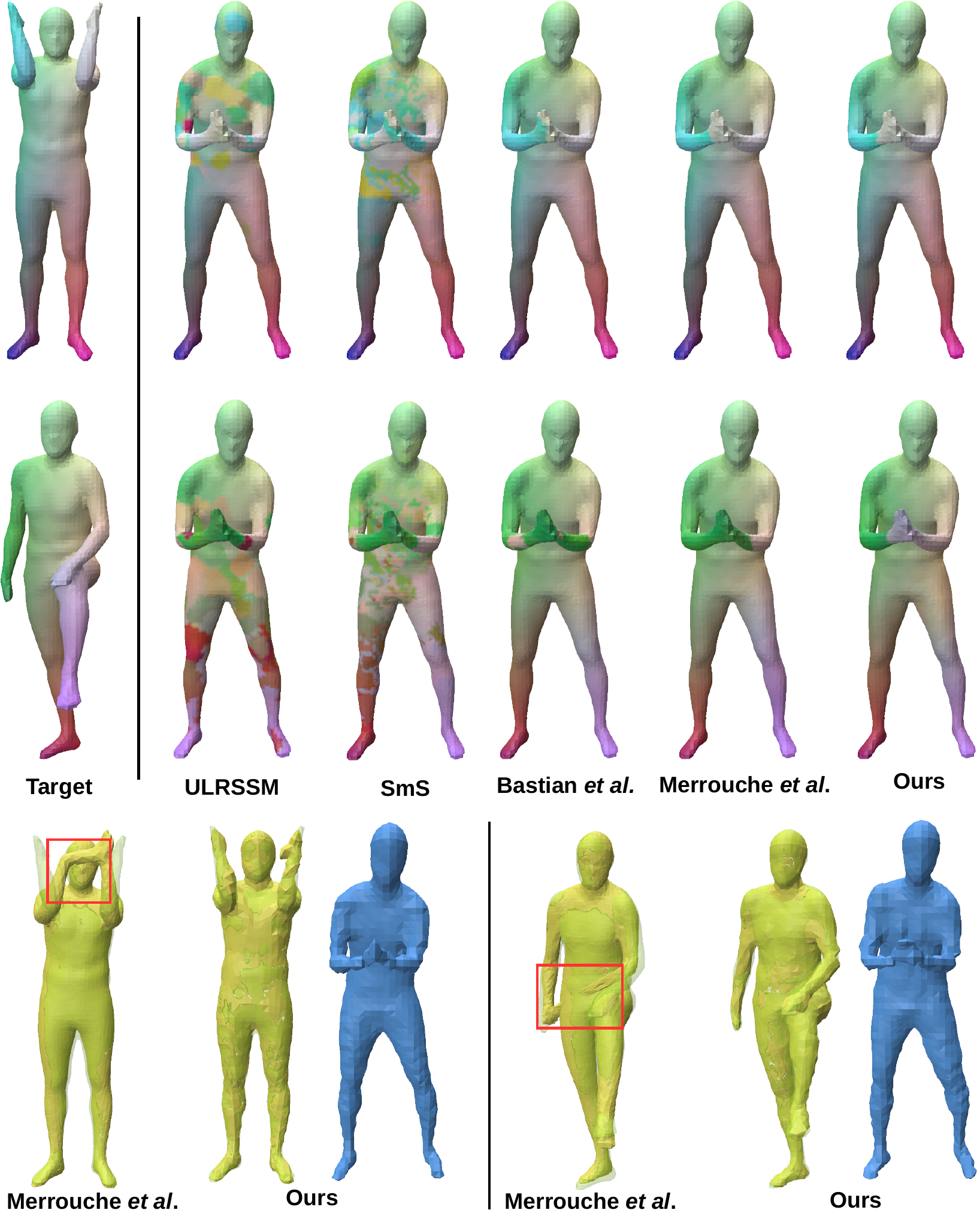}
    \caption{Correspondence results on ExtFAUST (top and second row). Colors are defined on the target shape and transferred to the source shape with correspondences output by each method. The last row shows the 3D alignments found by the approach of Merrouche~\etal~\cite{Merrouche_2023_BMVC} and ours. It also shows in blue the template found by our method.}
    \label{fig:supp_contact}
\end{figure}

\begin{figure}[h]
    \centering
    \includegraphics[width=1.0\columnwidth]{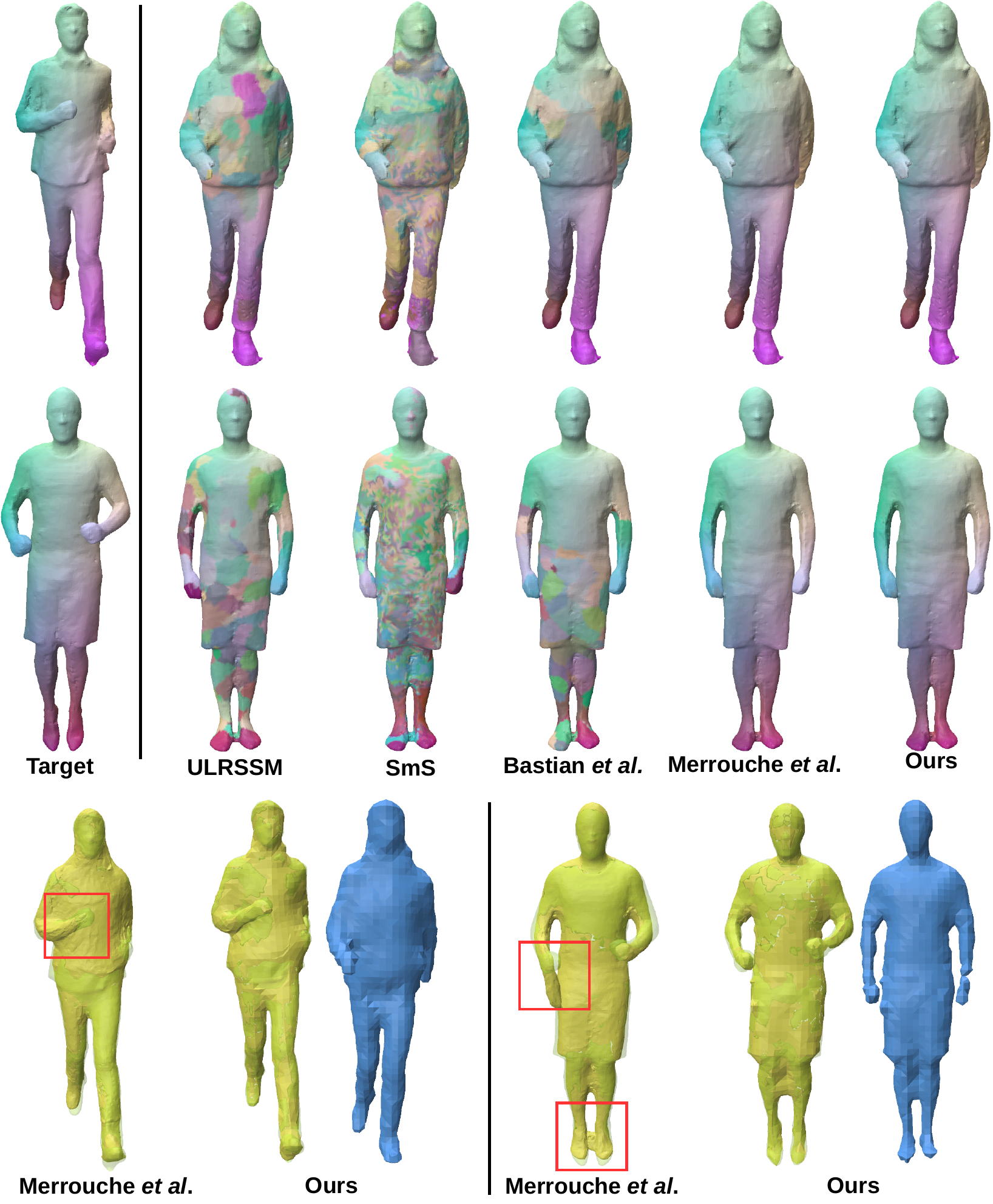}
    \caption{Correspondence results on 4DHumanOutfit (top and second row). Colors are defined on the target shape and transferred to the source shape with correspondences output by each method. The last row shows the 3D alignments found by the approach of Merrouche~\etal~\cite{Merrouche_2023_BMVC} and ours. It also shows in blue the template found by our method.}
    \label{fig:supp_4dho}
\end{figure}

We provide additional qualitative comparisons with FM based methods ULRSSM~\cite{cao2023unsupervised}, SmS~\cite{cao2024spectral}, Bastian~\etal~\cite{bastianxie2024hybrid} (Hybrid ULRSSM) and deformation-guided method Merrouche~\etal~\cite{Merrouche_2023_BMVC} on ExtFAUST~\cite{basset2021neural, Merrouche_2023_BMVC} in Fig.~\ref{fig:supp_contact} and 4DHumanOutfit~\cite{armando20234dhumanoutfit} in Fig.~\ref{fig:supp_4dho}. In each figure the top and second rows give the correspondence results where colors are defined on the target mesh and transferred to the source mesh using the correspondences output by each method, and the last row compares our alignments with those output by the approach of Merrouche~\etal~and shows in blue the template found by our method.

\subsection{Failure Cases}

\begin{figure}[h]
    \centering
    \includegraphics[width=1.0\columnwidth]{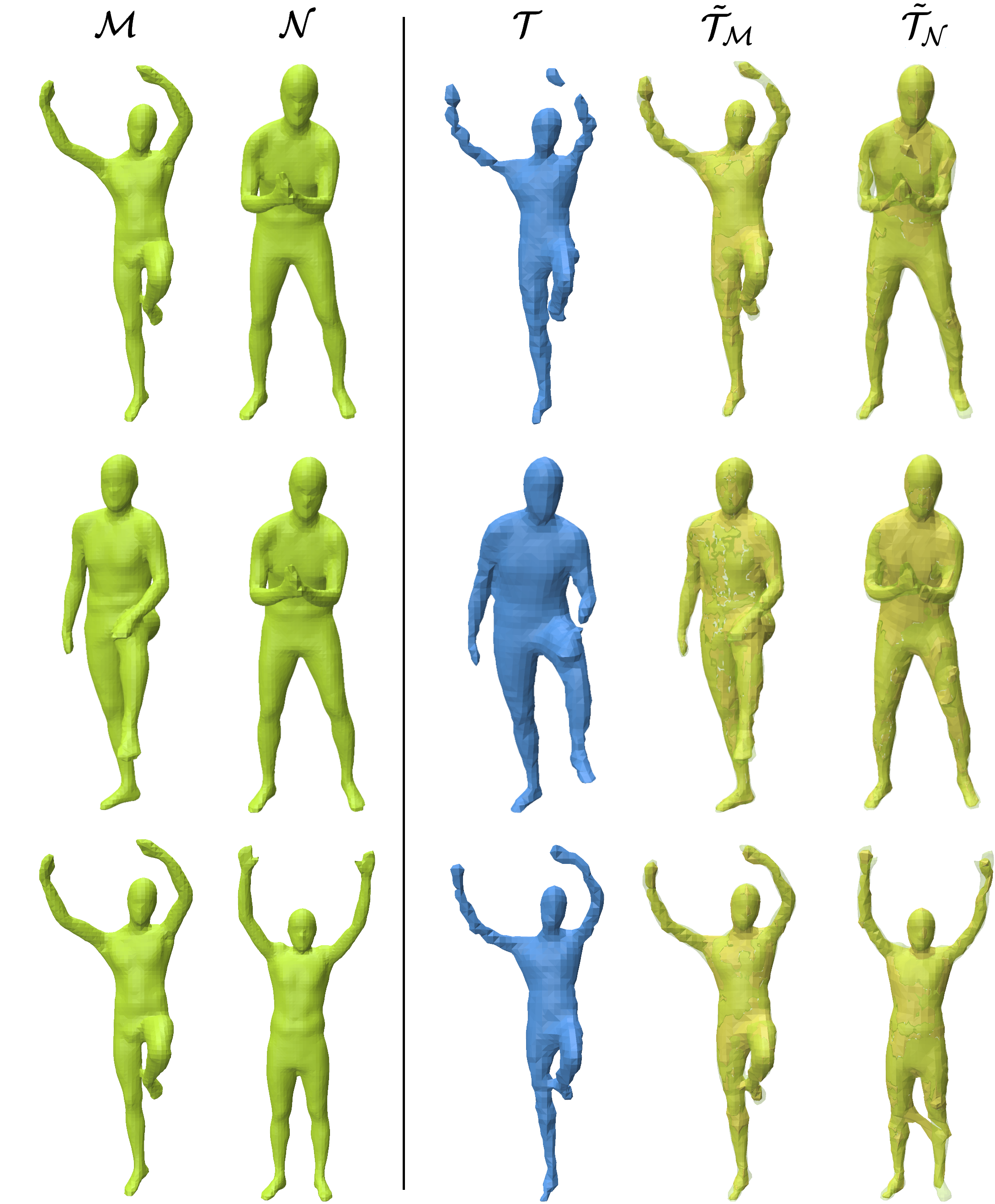}
    \caption{Examples where the topology optimisation strategy gets stuck in local minima.}
    \label{fig:failures_cases}
\end{figure}

As the topology optimisation leverages a greedy hill climbing like strategy, it can get stuck in local topology minima. Fig.~\ref{fig:failures_cases} gives examples, it shows the initial shapes, the template that was found and its alignments with the initial shapes, respectively. The template is initialised as $\M$. The topology optimisation creates a disconnected component in the first example, performs a sub-optimal cut (cutting the arm at the wrist and leaving the hand attached) in the second one and fails to cut the ankle in the last one.

\section{Implementation Details}
\label{sec:impelm_details}
\subsection{Neural Field Fitting}
\label{sec:fitting_supp}

The individual loss terms in the neural field fitting objective (Eq.~\ref{fitting_loss}) are implemented as follows:

{
\footnotesize
\begin{equation}
\label{gt_sdf_loss}
    l_{SDF}(\mathcal{F}, \Sigma) = \frac{1}{S} \sum\limits_{j = 1}^{S} (|S_{\Sigma}(\operatorname{tri}(\mathcal{F}, x_j))   - gt_{sdf}(x_j) |)
\end{equation}

\begin{equation}
\label{eikonal_loss}
    l_{eikonal}(\mathcal{F}, \Sigma) = \frac{1}{S} \sum\limits_{j = 1}^{S} (\|\|\nabla_{x_j}S_{\Sigma}(\operatorname{tri}(\mathcal{F} , x_j))\|_2   - 1\|_2^2)
\end{equation}

\begin{equation}
\label{smoothing_loss}
    l_{smoothing}(\mathcal{F}) = \| \mathcal{F} - \mathcal{G}_{k, \sigma} * \mathcal{F} \|_2^2
\end{equation}

\begin{equation}
\label{reg_loss}
    l_{reg}(\mathcal{F}, \Sigma) = \| \mathcal{F} \|_2^2 + \| \Sigma \|_2^2
\end{equation}
}

where $(x_j \in B)_{j \in \{1,...,S\}}$ are $S$ points sampled in the bounding box $B$ of $\T$, $gt_{sdf}(x_j)$ is the ground truth SDF value at point $x_j$, $\mathcal{G}_{k, \sigma}$ is the Gaussian kernel with size $k$ and standard deviation $\sigma$ and $*$ the convolution operation. 

The bounding box $B$'s extents are fixed to $(0.5,0.5,0.5)$ and $(-0.5,-0.5-,0.5)$. We use $|S|=500000$ samples, with a ratio of $20\%$ points sampled uniformly in $B$ and $80\%$ sampled within a distance of $ \pm 0.05$ to $\T$. For Gaussian smoothing, we fix $k=3$ and $\sigma=1$. We fix the loss term weights to $\lambda_1=1, \lambda_2=1e-1, \lambda_3=5e4, \lambda_4=1e-4$. The neural field is optimised using Adam~\cite{kingma2014adam} for $1500$ iterations; the learning rate starts at 1e-3 and is halved at epochs 300, 600, 900, and 1200. Neural field fitting takes approximately $4mn$ on an NVIDIA RTX A6000 GPU.

\begin{figure}[h]
    \centering
    \includegraphics[width=0.4\columnwidth]{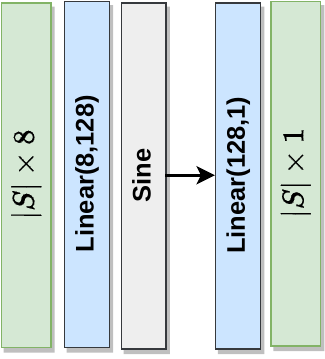}
    \caption{$S_{\Sigma}$'s architecture.}
    \label{fig:mlp}
\end{figure}

In our experiments we use a feature volume $\mathcal{F} \in \mathbb{R}^{(128 \times 128 \times 128) \times 8}$ of resolution $128$ and feature size $8$. $S_\Sigma$'s architecture is shown in Fig.~\ref{fig:mlp}.

\subsection{Patch Decomposition}

To obtain the surface patches $(P_k)_{1 \leq k \leq L}$, we employ a furthest point sampling like strategy using geodesic distances, similarly to~\cite{Merrouche_2023_BMVC}. In all our experiments, we use $L=200$ surface patches.

\subsection{Association Model}

\begin{figure}[h]
    \centering
    \includegraphics[width=1\columnwidth]{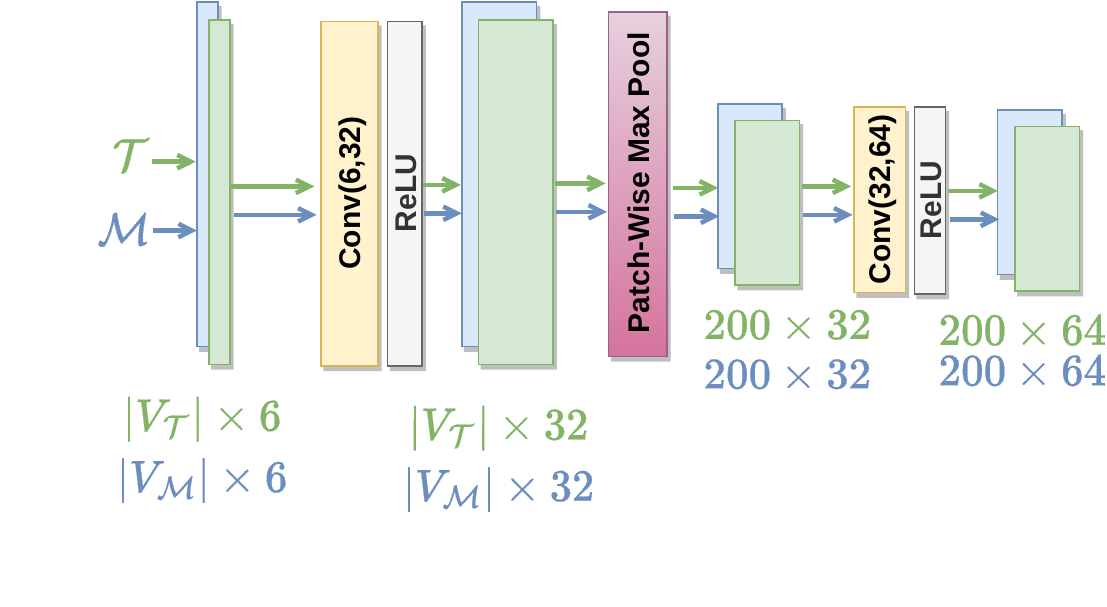}
    \caption{$A_{\Theta}$'s architecture.}
    \label{fig:a_theta}
\end{figure}

The association network $A_{\Theta}$'s architecture is shown in Fig.~\ref{fig:a_theta}, it uses the graph convolution operator from~\cite{verma2018feastnet}.

The individual loss term weights in Eq.~\ref{eq:association_loss} are fixed to $\gamma_1=1e4, \gamma_2=5e3$.

\subsection{Deformation Model}

\begin{figure}[h]
    \centering
    \includegraphics[width=1.0\columnwidth]{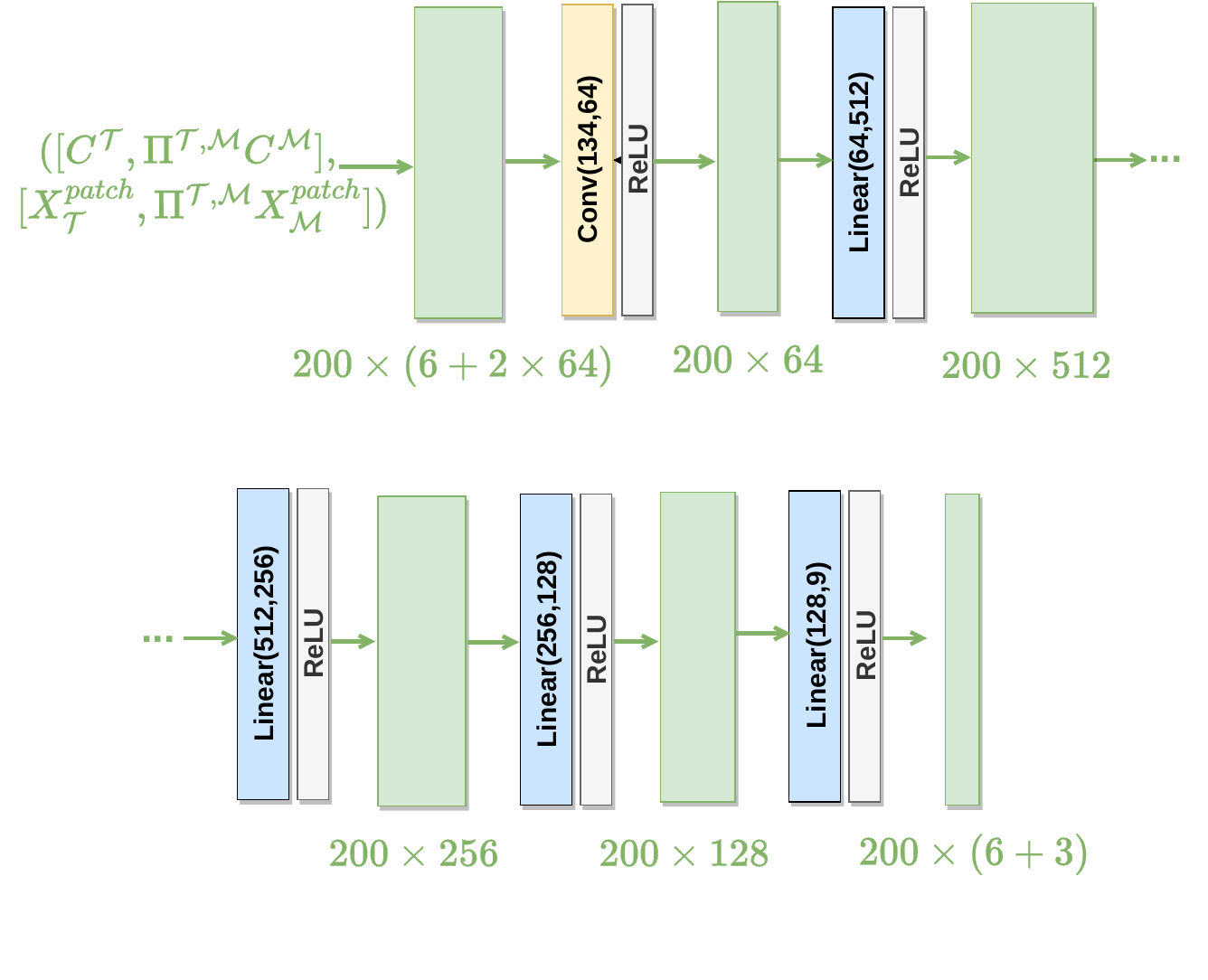}
    \caption{$D_{\Psi}$'s deformation decoder architecture. ``...'' indicate continued connection.}
    \label{fig:d_psi}
\end{figure}

The deformation network's architecture is composed of two sub-modules. A feature extractor with an architecture similar to $A_{\Theta}$'s and a deformation decoder. Denoting $X^{patch}_{\T}$ and $X^{patch}_{\M}$ the features extracted by this feature extractor for shapes $\T$ and $\M$ respectively, the deformation network's architecture can be seen in Fig.~\ref{fig:d_psi}. For each surface patch, it outputs $3$ translation parameters and $6$ rotation parameters as the $6D$ parameterisation of rotations is used~\cite{zhou2019continuity}.

The individual loss term weights in Eq.~\ref{eq:def_loss} are fixed to $\beta_1=1e4, \beta_2=5e2, \beta_3=1e6$. For the silhouette loss, we use $|\Omega|=50$ viewpoints sampled uniformly on a sphere of radius $1/2$ centered at the bounding box $B$'s center. For each viewpoint, we obtain silhouette images of resolutions $H\times W = 256\times256$ smoothed with a Gaussian kernel of size $k=5$ and standard deviation $\sigma=1$.

\subsection{Topology Update Steps}

We fix the individual loss term weights in Eq.~\ref{eq:top_def_loss} to $\alpha_1=1, \alpha_2=5e2, \alpha_3=2, \alpha_4=5e4$ and use a tolerance parameter of $\tau=2e-4$ for $l_{min}$.

\subsection{Optimisation}
\label{sec:opt_details}
Our detailed topology-adaptive deformation-guided mesh matching optimisation is shown in Alg.~\ref{algo_opt}. 

Note that we use the same networks $A_{\Theta}$ and $D_{\Psi}$ to obtain associations and deformations between $\T$ and $\M$, $\N$. 

Also note that during topology update steps (line 28 of Alg.~\ref{algo_opt}), we only leverage $l_{topo}(\N, \tilde{\T^i}_{\N})$. This is since $\T^{0}$ corresponds to $\M$, hence $l_{min}(\T^i, \T^0, \tau)$ will encourage $\T^i$ to stay geometrically close to $\M$. Further, note that we perform multiple inner topology gradient descent steps $k$ for each iteration $i$ (line 27 of Alg.~\ref{algo_opt}), after which estimating new associations and deformations with the current estimates of $\Theta$ and $\Psi$ is necessary  (lines 30 and 31 of Alg.~\ref{algo_opt}). We do this to accelerate the optimisation, counting on the graph GNN parametrisation of $A_{\Theta}$ and $D_{\Psi}$'s robustness.

To perform gradient descent on $l_{assos}$, $l_{def}$ and $l_{topo}$ (lines 11, 20 and 28 of Alg.~\ref{algo_opt} respectively), we use the Adam optimiser~\cite{kingma2014adam} with learning rates $\eta_1=1e-3$,$\eta_2=1e-3$ and $\eta_3=1e-4$ respectively. To obtain $\nabla_{(\mathcal{F}^{(i, k)}, \Sigma^{(i, k)})} \big(\alpha_1 {l}_{sil}(\N, \tilde{\T^i}_{\N})  + \alpha_2 l_{min}(\T^i, \T^0, \tau)\big)$ from $\nabla_{V_{\T^i}} \big(\alpha_1 {l}_{sil}(\N, \tilde{\T^i}_{\N})  + \alpha_2 l_{min}(\T^i, \T^0, \tau)\big)$ in line 28 of Alg.~\ref{algo_opt}, we use the equation of MeshSDF~\cite{remelli2020meshsdf}.

We fix the number of iterations $n$ to $500$ for the SMAL, SCAPE and FAUST datasets and $2000$ for the ExtFAUST and 4DHumanOutfit datasets. For all datasets we fix the number of inner iterations to $n_1=1, n_2=10, n_3=5$. We start topology update steps after $n_{top}=300$ iterations, to ensure that we have found good initial associations and deformations. After $n_{top}$, a topology update step is performed every $K=10$ iterations which corresponds to the period $K$ of the periodic weighting scheme detailed in the next paragraph.

\begin{algorithm*}
\caption{Topology-Adaptive Deformation-Guided Mesh Matching}
\label{algo_opt}
\begin{algorithmic}[1]
\State \textbf{Input:} Meshes $\mathcal{M}$, $\mathcal{N}$

 \State \textbf{Initialize:} \( \tilde{\T^0}_{\M}, \tilde{\T^0}_{\N}, \T^0 \gets \M, \M, \M \)

 \State Fit neural field of parameters ($\mathcal{F}^{0}$, $\Sigma^0$) to $\T^0$.

\For{each outer iteration \( i = 0, \dots, n \)}
    \State Obtain patch segmentation for $\T^i$, $\M$ and $\N$

    \State \textbf{// Optimize \( \Theta^i \)}: bijective associations search
    \State $\Theta^{(i, 0)} \gets \Theta^i$
    \For{step \( l = 0, \dots, n_1 \)}
        \State \(F^{patch}_{\T^i}, F^{patch}_{\M}, F^{patch}_{\N} \gets A_{\Theta^{(i, l)}}(\T^i), A_{\Theta^{(i, l)}}(\M), A_{\Theta^{(i, l)}}(\N)\)
        \State \(\Pi^{\T^i, \M}, \Pi^{\T^i, \N} \gets \sinkhorn(\cosim(F^{patch}_{\T^i}, F^{patch}_{\M})), \sinkhorn(\cosim(F^{patch}_{\T^i}, F^{patch}_{\N})) \)
        \State \( \Theta^{(i, l+1)} \gets \Theta^{(i,l)} - \eta_1 \nabla_{\Theta^{(i, l)}}  \big(l_{assos}(\T^i, \M)+l_{assos}(\T^i, \N)\big) \)
    \EndFor
    \State $\Theta^{i+1} \gets \Theta^{(i, n_1)}$
    \State \textbf{// Optimize \( \Psi^i \)}: ARAP deformation search
    \State $\Psi^{(i, 0)} \gets \Psi^i$
    \For{step \( j = 1, \dots, n_2 \)}
        \State \( R_{\T^i, \M}, U_{\T^i, \M} \gets D_{\Psi^{(i,j)}}(\T^i, \M, \Pi^{\T^i, \M})\)
        \State \( R_{\T^i, \N}, U_{\T^i, \N} \gets D_{\Psi^{(i,j)}}(\T^i, \N, \Pi^{\T^i, \N})\)
        \State \( \tilde{\T^i}_{\M}, \tilde{\T^i}_{\N} \gets \apdef(\T^i, R_{\T^i, \M}, U_{\T^i, \M}), \apdef(\T^i, R_{\T^i, \N}, U_{\T^i, \N})\) 
        \State \( \Psi^{(i,j+1)} \gets \Psi^{(i,j)} - \eta_2 \nabla_{\Psi^{(i,j)}} \big(l_{def} (\T^i, \M, \Pi^{\T^i, \M}) + l_{def} (\T^i, \N, \Pi^{\T^i, \N})\big) \)
    \EndFor
    \State $\Psi^{i+1} \gets \Psi^{(i, n_2)}$

    \State \textbf{// Optimize \( (\mathcal{F}^i, \Sigma^i) \)}: topology adaptation
    \If{\((i \geq n_{top}) \textit{ and } (i \bmod K = 0)\)}
        \If{\big($l_{sil}(\M, \tilde{\T^{i}}_{\M}) + l_{sil}(\N, \tilde{\T^{i}}_{\N})<l_{sil}(\M, \tilde{\T^{i-1}}_{\M}) + l_{sil}(\N, \tilde{\T^{i-1}}_{\N})$\big) \textit{or} $(i=n_{top})$}:
            \State $(\mathcal{F}^{(i, 0)}, \Sigma^{(i, 0)}) \gets (\mathcal{F}^i, \Sigma^i$) 
            \For{step \( k = 0, \dots, n_3 \)}
                
                \State \( (\mathcal{F}^{(i, k+1)}, \Sigma^{(i, k+1)})  \gets (\mathcal{F}^{(i, k)}, \Sigma^{(i, k)}) - \eta_3 \nabla_{(\mathcal{F}^{(i, k)}, \Sigma^{(i, k)})} l_{topo}(\N, \tilde{\T^i}_{\N})\)
                \State \(  \T^i \gets \mcubes(\mathcal{F}^{(i, k+1)}, S_{\Sigma^{(i, k+1)}}) \)
                \State Get patch segmentation for $\T^i$
                \State Get new $\Pi^{\T^i, \M}, \Pi^{\T^i, \N}$ (as in line~10) and new $\tilde{\T^i}_{\M}, \tilde{\T^i}_{\N}$ (as in line~19) using $\Theta^{i+1},\Psi^{i+1}$ 
            \EndFor
            \State $(\mathcal{F}^{i+1}, \Sigma^{i+1}) \gets (\mathcal{F}^{(i, n_3)}, \Sigma^{(i, n_3)})$
        \Else 
            \State $(\mathcal{F}^{i+1}, \Sigma^{i+1}), \Theta^{i+1}, \Psi^{i+1} \gets (\mathcal{F}^{i-1}, \Sigma^{i-1}), \Theta^i, \Psi^{i}$
        \EndIf
    \Else  
        \State $(\mathcal{F}^{i+1}, \Sigma^{i+1}) \gets (\mathcal{F}^{i}, \Sigma^{i})$
    \EndIf
    \State \(  \T^{i+1} \gets \mcubes(\mathcal{F}^{i+1}, S_{\Sigma^{i+1}}) \)
\EndFor
\State $\T, \tilde{\T}_{\M}, \tilde{\T}_{\N} \gets \T^n, \tilde{\T^n}_{\M}, \tilde{\T^n}_{\N}$
\State Compute $\phi : V_{\M} \to V_{\N}$, as $\phi(x) = \arg\min_{w \in V_{\N}} \left\| w - C\left( \arg\min_{v \in V_{\tilde{\T}_{\M}}} \| x - v \|_2 \right) \right\|_2 \forall x \in V_{\M}$, where $C$ maps the $i^{th}$ vertex of $\tilde{\T}_{\M}$ to the $i^{th}$ vertex of $\tilde{\T}_{\N}$.
\State \textbf{Output:$\T, \tilde{\T}_{\M}, \tilde{\T}_{\N}, \phi$} 
\end{algorithmic}
\end{algorithm*}

\noindent\textbf{Periodic Weighting Scheme}
We alternate between optimising for bijective associations, ARAP deformations under the latter and topology updates. To benefit the most from the synergy of these steps, we use a periodic weighting scheme $w_K(i)$ throughout the optimisation iterations \( i = 0, \dots, n \).  Equations~\ref{eq:association_loss} and ~\ref{eq:def_loss} of the main paper become:

{
\footnotesize
\begin{align}
    l_{assos}(\T, \M) &=  (1-w_K(i)) \gamma_1 l_{perm}(\Pi^{\T, \M}) \\
    &+ w_K(i) \gamma_2 l_{match}(\Pi^{\T, \M}, \tilde{\T}_{\M}, \M) \nonumber
\end{align}
}

{
\footnotesize
\begin{align}
    l_{def}(\T, \M, \Pi^{\T, \M}) &=  (1-w_K(i)) \beta_1 l_{match}(\Pi^{\T, \M}, \tilde{\T}_{\M}, \M) \\
    &+ w_K(i)\beta_2 l_{sil}(\M, \tilde{\T}_{\M}) + \beta_3 l_{rig}(R_{\T, \M}, U_{\T, \M}) \nonumber
\end{align}
}

where $w_K(i)$ is written by the equation:

{
\footnotesize
\begin{equation}
     w_K(i) = 1/2 \left( 1 + \cos \left( 2 \pi i / K \right) \right)
\end{equation}
}

\begin{figure}[h]
    \centering
    \includegraphics[width=1.0\columnwidth]{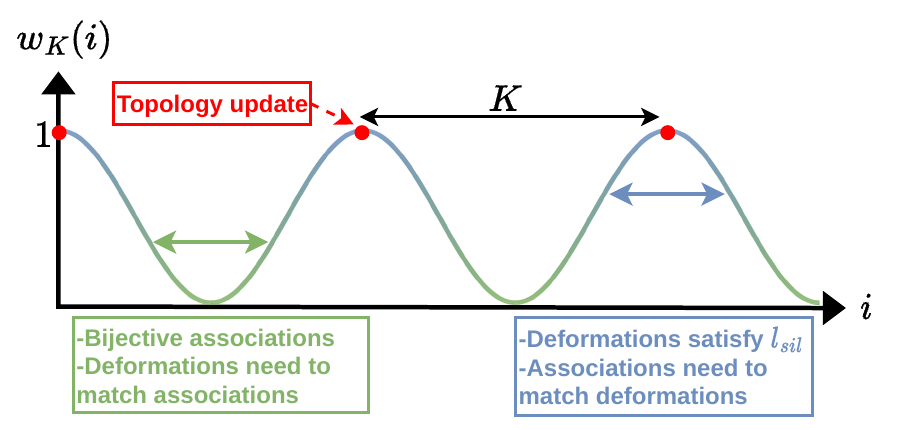}
    \caption{Periodic weighting scheme $w_K(i)$.}
    \label{fig:periodic_weights}
\end{figure}

As shown in Fig.~\ref{fig:periodic_weights}, when $w_K(i)$ is close to $0$, corresponding to green, we search for bijective associations without deformation guidance. Adversely, deformations follow the bijective associations and the silhouette loss is relaxed.

When $w_K(i)$ is close to $1$, corresponding to blue, associations are deformation guided and the bijectivity constraint is relaxed. Adversely, deformations are assessed with the silhouette loss, while able to diverge from the associations.

Topological update steps, corresponding to the red dots, are only conducted when the alignment, as assessed with the silhouette loss, is at its best,~\ie~ $w_K(i)=1$. 

The optimisation takes approximately $14mn$ for pairs from FAUST, SCAPE and SMAL and approximately $54mn$ for pairs from ExtFAUST and 4DHumanOutfit. We experiment on an NVIDIA RTX A6000 GPU.

\end{document}